\documentclass{article}

 \usepackage[dblblindworkshop, final]{neurips_2025}

\workshoptitle{Data on the Brain \& Mind}

\usepackage[utf8]{inputenc} 
\usepackage[T1]{fontenc}    
\usepackage{hyperref}       
\usepackage{url}            
\usepackage{booktabs}       
\usepackage{amsfonts}       
\usepackage{nicefrac}       
\usepackage{microtype}      
\usepackage{xcolor}         
\usepackage{amsmath}
\usepackage{dsfont}

\hypersetup{
  colorlinks=true,
  linkcolor=blue,
  citecolor=blue,
  urlcolor=blue
}

\usepackage{dsfont} 

\usepackage{graphicx}
\graphicspath{{figures/}}
\usepackage{float}
\usepackage{placeins}

\newcommand{\pnpl}{PNPL~\raisebox{-0.2ex}{\includegraphics[height=1.6ex]{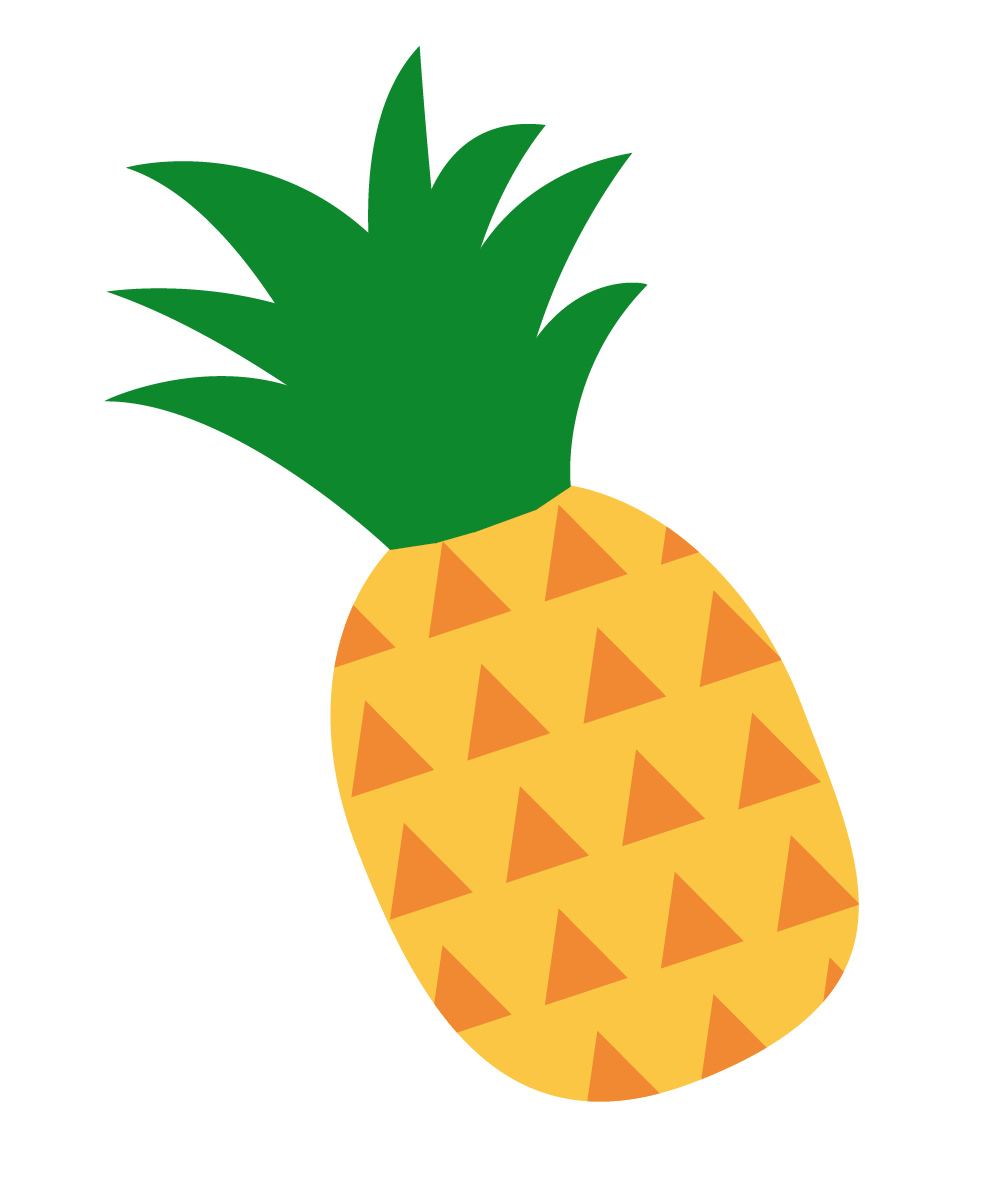}}}

\title{Elementary, My Dear Watson: Non-Invasive Neural Keyword Spotting in the LibriBrain Dataset}

\author{%
  Gereon Elvers \\
  \pnpl\\
  Department of Engineering Science\\
  University of Oxford, UK \\
  \texttt{gereon.elvers@tum.de} \\
  \And
  Gilad Landau \\
  \pnpl\\
  Department of Engineering Science\\
  University of Oxford, UK \\
  \texttt{gilad@robots.ox.ac.uk} \\
  \And
  Oiwi Parker Jones \\
  \pnpl\\
  Department of Engineering Science\\
  University of Oxford, UK \\
  \texttt{oiwi@robots.ox.ac.uk} \\
}

\begin{document}

\maketitle

\vspace{-.075cm} 

\begin{abstract}
  Non-invasive brain-computer interfaces (BCIs) are beginning to benefit from large, public benchmarks. However, current benchmarks target relatively simple, foundational tasks like Speech Detection and Phoneme Classification, while application-ready results on tasks like Brain-to-Text remain elusive. We propose Keyword Spotting (KWS) as a practically applicable, privacy-aware intermediate task. Using the deep 52-hour, within-subject LibriBrain corpus, we provide standardized train/validation/test splits for reproducible benchmarking, and adopt an evaluation protocol tailored to extreme class imbalance. Concretely, we use area under the precision-recall curve (AUPRC) as a robust evaluation metric, complemented by false alarms per hour (FA/h) at fixed recall to capture user-facing trade-offs. To simplify deployment and further experimentation within the research community, 
  we are releasing an updated version of the \texttt{pnpl} library with word-level dataloaders and Colab-ready tutorials. As an initial reference model, we present a compact 1-D Conv/ResNet baseline with focal loss and top-$k$ pooling that is trainable on a single consumer-class GPU. The reference model achieves $\sim$$13\times$ the permutation-baseline AUPRC on held-out sessions, demonstrating the viability of the task. Exploratory analyses reveal: (i) predictable within-subject scaling—performance improves log-linearly with more training hours—and (ii) the existence of word-level factors (frequency and duration) that systematically modulate detectability.
\end{abstract}

\vspace{-.075cm} 

\begin{figure}[!ht]
\centering
\includegraphics[width=\linewidth]{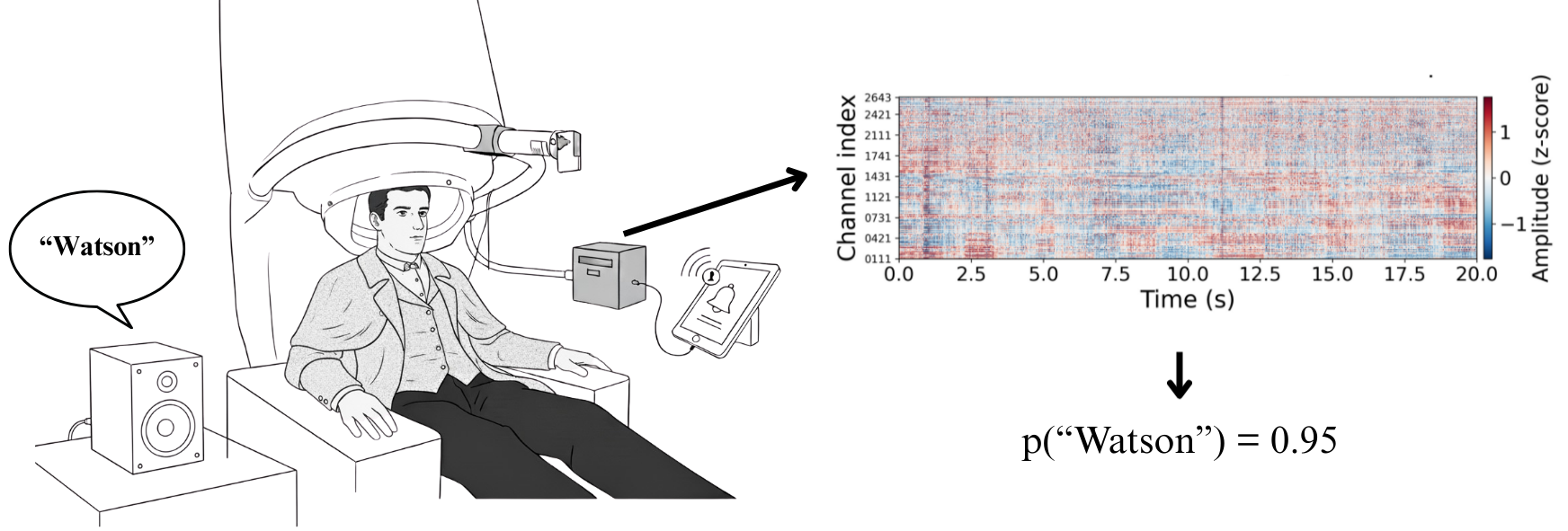}
\caption{KWS task setup. The participant listens to a Sherlock Holmes audiobook while the model spots a chosen keyword (e.g., ``Watson'') from MEG signals.}
\label{fig:task-graphic}
\end{figure}

\section{Introduction}
The arrival of large and readily available datasets has begun to supply non-invasive brain-computer-interface (BCI) research with the kind of ``common yard-stick'' that ImageNet \citep{russakovsky2015imagenet} provided for computer vision. Among current non-invasive datasets for decoding speech, LibriBrain \citep{ozdogan2025dataset} is the \textit{deepest} (i.e., \textit{largest within-subject}) with 52 hours of magnetoencephalography (MEG) recorded from a single participant. This dataset forms the foundation for the 2025 PNPL Competition \citep{landau2025competition}, an open machine-learning competition that has catalyzed progress on two foundational decoding tasks: Speech Detection and Phoneme Classification. Progress on these tasks can be seen by looking at the online leaderboards (\url{https://libribrain.com/}). For example, in just two months F1 macro scores on the Speech Detection task advanced rapidly from about 68\% to a new state-of-the-art on the (extended) public leaderboard of 96\%. Our aim in this paper is to build on this success by introducing a new standardized task: Word Detection (a.k.a.~Keyword Spotting). Substantively, we provide the same kinds of supporting infrastructure for this task (e.g., data loader, reference model, reproducible metrics, public leaderboard) which led directly to accelerated improvements in Speech Detection and Phoneme Classification. 

As \citet{landau2025competition} explain, the two tasks in the 2025 PNPL Competition were selected for their simplicity. Over time, the idea was to increase the complexity, and utility, of benchmark decoding tasks. Keyword Spotting (KWS) is an exciting landmark as it represents the first decoding task on this curriculum with practical utility for BCIs. In the established domain of voice computing, Keyword Spotting is commonly used to detect ``wake words'' (e.g., ``Hey Siri'', ``Alexa'', ``OK Google''). Wake words like these can be used to indicate that subsequent speech should be interpreted as a command, or they can be used themselves as commands. In the emerging domain of \textit{brain} computing, even a single wake word (e.g., ``help'') could be profoundly meaningful to someone with severe paralysis. Only slightly further along the curriculum, a small set of working keywords (e.g., ``hungry'', ``tired'', ``thirsty'', ``toilet'', ``pain'') would transform their quality of life. The benchmark task established in this paper is intended, ultimately, to lead to such an outcome. For clarity, and to contrast the use of acoustic inputs, we use the term \textit{Neural Keyword Spotting} 
to denote Keyword Spotting from continuous brain data, a task represented schematically in Figure \ref{fig:task-graphic}.

\section{Related Work}

\subsection{Existing Tasks for Non-Invasive Speech Decoding}

As mentioned in the introduction, the introduction of a rigorous benchmark a recent development in non-invasive speech decoding. Here a \textit{benchmark} is a set of resources used by the community to measure progress. Benchmarks go beyond typical published work by including the following to support the measurement of progress. 
\begin{itemize}
    \item \textbf{Standardized data}: Publicly available, with well-defined train and holdout splits. 
    \item \textbf{Evaluation metrics}: Agreed measure of success with public leaderboard to track progress.
    \item \textbf{Reference model}: Reproducible implementation with open weights and training code.
\end{itemize}
Prior to the release of LibriBrain \citep{ozdogan2025dataset}, together with a standard Python library (\texttt{pnpl}) for loading predefined data splits \citep{landau2025competition}, a number of open datasets \citep[e.g.,][]{schoffelen2019mous,armeni2022narratives,gwilliams2023megmasc} were starting to reappear across large-scale studies \citep[e.g.,][]{defossez2023nmi, dascoli2024words, ridge2024domainshift, jayalath2025icml}. 
However, data splits were not generally replicated making it difficult to compare methods. The same model architectures and weights were neither generally shared nor used as baselines and there were no public leaderboards. 

A rich set of speech decoding tasks have nonetheless emerged. These include the following tasks.

\textbf{Brain-to-Text (B2T)} takes variable-length neural sequences (e.g., EEG, MEG, fMRI) as input and outputs text transcripts, typically evaluated with word error rate (WER) or semantic similarity metrics such as the BERTScore \citep{bertscore}. B2T is the analogue of ASR and represents a long-term goal, though its difficulty has motivated the development of what we call intermediate tasks, which lie between more foundational tasks like Speech Detection and full B2T. Non-invasive B2T has been explored with EEG and MEG \citep{duan2023dewave, jo2024eegtotext, yang2024neuspeech, yang2024mad, yang2024neugpt}, with semantic metrics often reported in place of WER, although recent work shows that competitive WERs are beginning to be achievable non-invasively \citep{jayalath2025unlocking}. In fMRI, the coarse temporal resolution makes word-level alignment unlikely, though remarkable paraphrases have been produced which retain some semantic similarities to the ground truth speech \citep[e.g.,][]{tang2023semantic}.


\textbf{Word Classification} uses fixed-length neural segments aligned to individual words, producing categorical labels from a closed vocabulary. A number of recent works have focused on vocabularies of 250 words \citep{dascoli2024words, ozdogan2025dataset, jayalath2025unlocking}, though recent models can also impute out-of-vocabulary items with an external LLM \citep[e.g.,][]{anthropic2025claude37} if the predicted word is “unknown” \citep{jayalath2025unlocking}.


\textbf{Phoneme Classification} operates on shorter neural segments aligned to phonemes, predicting categorical labels over the phoneme inventory \citep{ozdogan2025dataset, landau2025competition}. Relatedly, Phonetic Feature Classification outputs binary labels for broader phonological features such as voicing. Phonetic features group together multiple phoneme classes (e.g., voiced /b, v, z/ vs. unvoiced /p, f, s/) and can therefore be more data-efficient \citep{gwilliams2022phonseq, jayalath2025icml}.


\textbf{Segment Identification} is a matching task. Given paired speech and brain data (e.g., cut continuous data into 3 second segments), the task is to correctly match audio and brain segments \citep{defossez2023nmi, tang2023semantic}. This task is only applicable when speech and brain data are temporally aligned, limiting its utility for BCIs.


\textbf{Speech Detection} works on potentially open-ended neural recording. The aim is to identify when subjects were processing speech. The use of the term \textit{processing} here is deliberate, as subjects could for example be listening to speech \citep{ozdogan2025dataset, landau2025competition} or speaking aloud \citep{Dash2020NeuroVAD}. There is a contrast between Speech Detection and Classification, though it is perhaps subtle. In Speech Classification, fixed-duration inputs are assigned to a class (e.g., speech or non-speech) \citep{jayalath2025icml}. Speech Classification models can be repurposed for Detection by applying them in sliding windows; but the task definitions remain formally distinct.

\subsection{Keyword Spotting}
Keyword spotting in the traditional audio domain (also referred to as wake-word detection) is a mature, highly-imbalanced detection problem optimizsed for very low false-alarm (FA) rates at fixed recall. Early small-footprint CNN and CRNN systems established the modern operating regime (e.g., 0.5 FA/h at acceptable FRR) under tight on-device constraints 
\citep[e.g., ][]{sainath2015cnn,arik2017crnn}. 
Large benchmarks like Speech Commands \citep{warden2018speechcommands}) and efficient architectures like MatchboxNet \citep{matchboxnet2020} and Keyword Transformer \citep{howard2021kwt} further drove accuracy/latency trade-offs for embedded devices, while industrial deployments (e.g., Apple's ``Hey Siri'') codified evaluation practices around FA/h and user-centric thresholds \citep{apple2017heysiri}.

On the invasive brain side, \citet{milsap2019keyword} introduced neural KWS with ECoG, showing low-latency, high-specificity detection using matched-filter templates spanning motor and auditory speech representations.
Recent intracortical studies push to large-vocabulary online decoding and inner-speech control, but their goals (continuous B2T, WER/CER) and signal quality differ materially from non-invasive KWS \citep{willett2023nature,metzger2023avatar,kunz2025inner}. 


Non-invasive technologies (EEG/MEG) have dramatic benefits over surgical implants in terms of safety and scalability. The application of keyword spotting is motivated by two converging strands. First, segment identification decoders trained to predict self-supervised speech representations from brain signals reliably retrieve the matching few-second stimulus among large candidate sets and generalise across participants 
- evidence that non-invasive signals carry phonetic/lexical detail at the granularity needed for lexical identification \citep{defossez2023nmi,dascoli2024words}.

Second, converging MEG/EEG results show sensitivity to phoneme sequence structure and higher-level linguistic content, and recent deep models capture meaningful portions of the speech-to-language transform in these signals \citep{gwilliams2022phonseq,tezcan2023tradeoff,desai2021generalizable}.
Against this recent progress, LibriBrain allows testing whether the same brain-speech representations that enable segment retrieval also support lexical selectivity for pre-specified words in its long-form, naturalistic stories. Due to its larger scale, it also allows building on prior EEG-based KWS pilots, which have largely remained at small-lexicon trialwise classification/onset detection \citep{sakthi2021keyword}.

\section{Methods}
\subsection{Dataset}
The following summary closely follows the original LibriBrain description. For full details, see \citep{ozdogan2025dataset}. 
In brief, the dataset covers over 52 hours of within-subject MEG data recorded on a 306-channel MEGIN Triux\textsuperscript{\texttrademark} Neo system (102 magnetometers, 204 planar gradiometers). Recordings were acquired at 1\,kHz and minimally preprocessed (head-motion correction; Maxwell filter; 50/100\,Hz notch filter; 0.1-125\,Hz band-pass filter) before downsampling to 250\,Hz (4\,ms samples), yielding data of shape \(C\times T\) with \(C{=}306\). Each session is paired with an \texttt{events.tsv} file listing onset/duration (s) for speech, word, and phoneme segments, all produced by forced-alignment \citep{gentle} and then manually corrected.

The release spans 93 sessions (3{,}139\,min; 52.32\,h) with 466{,}230 word tokens (16{,}892 unique) and 1{,}511{,}732 phoneme tokens. See Figures \ref{fig:libribrain-overview} and \ref{fig:libribrain-phoneme-duration} for an overview of the dataset. Word frequencies are Zipfian, providing keywords across a wide base-rate spectrum (short, frequent function words vs.\ longer, rarer content/proper names). These properties suit event-referenced keyword detection with extreme class imbalance.

\subsection{Task Definition}
We cast neural keyword spotting (KWS) from MEG as an event-referenced detection task using LibriBrain word onsets \citep{ozdogan2025dataset}. This can be formalized as follows: First, we fix a small keyword set $\mathcal{V}$ (minimally $|\mathcal{V}|=1$). For each keyword $k\in\mathcal{V}$, let $d_{\max}(k)$ be the maximum duration of any instance of $k$ in the corpus. Given that we may want to extract brain recordings that start and end before and after audio event boundaries (e.g., because neural processing continues after the presentation of a stimulus), we can select fixed pre/post buffers $\beta^{-}\!\ge 0$ and $\beta^{+}\!\ge 0$. 
These offsets can then be used to define $D(\mathcal{V})$, which is the total window duration (in seconds) of neural data to extract around any keyword in $\mathcal{V}$:
\[
D(\mathcal{V}) = \beta^{-} + \max_{k \in \mathcal{V}} d_{\max}(k) + \beta^{+}.
\]

Concretely, for each word token with onset $t_i$ and string $s_i$, we extract a $306\times D(\mathcal{V})$ window starting at $t_i-\beta^{-}$. This guarantees that any instance of any $k\in\mathcal{V}$ fits fully inside the window while allowing for a longer window duration if further context can improve detection. The binary label is
\[
y_i \;=\; \mathds{1}\{\,s_i\in\mathcal{V}\,\}\in\{0,1\},
\]
where $\mathds{1}\{\cdot\}$ is the indicator function. 

In contrast to KWS, full-vocabulary Brain-to-Text aims to identify $w\in\mathcal{W}$ (hundreds of thousands of words), which introduces severe long-tail sparsity and requires calibrating thresholds across many classes. KWS is a practical, fixed-lexicon task: it asks only whether any member of a small, predefined set $\mathcal{V}$ occurred. This offers a simple, reliable trigger with clean control over latency and false alarms—useful both on its own (assistive or hands-free commands) and as a stepping stone towards richer decoders.

By default, we adopt LibriBrain's session-level train/val/test split \citep{ozdogan2025dataset}. For a chosen set of keywords $\mathcal{V}$, we verify that positives occur in both validation and test. If not, we replace them with the two sessions containing the most positives for $\mathcal{V}$. 
This ensures sufficient positive examples for reliable metric computation, particularly important given the extreme class imbalance in keyword detection. For a session $S$ containing word tokens with strings $\{s_i\}_{i=1}^{n_S}$, let
\[
c_S = \sum_{i=1}^{n_S} \mathds{1}\{s_i \in \mathcal{V}\}
\]
be the count of keyword instances in session $S$. Validation and test are set to the two sessions maximizing $c_S$.

\subsection{Metrics}
\label{subsec:metrics}
We use area under the precision-recall curve (AUPRC) as the primary metric. This is well suited to keyword spotting because its base rate equals the empirical prevalence of positives, so gains are easy to interpret. It also summarises the trade-off we actually care about under heavy imbalance—how many of the system's alarms are correct (precision) as we demand more coverage (recall). Finally, unlike alternative metrics (e.g., AUROC), AUPRC is not overly optimistic when precision is too low to be usable. We supplement AUPRC with additional metrics like AUROC to provide a comprehensive view.
For scenario-grounded reporting, we translate any validation-selected threshold (precision $P$, recall $R$) into hourly rates under an assumed event frequency $\lambda$ (keywords/hour):
\[
\mathrm{FA/h} = R\,\lambda\Big(\tfrac{1}{P}-1\Big),\qquad
\text{Misses/h} = \lambda(1-R),\qquad
\text{Detections/h} = \lambda R.
\]
We consider two illustrative use cases: assistive access ($\lambda\!\approx\!2$/h) and hands-free control ($\lambda\!\approx\!10$/h). Thresholds $\tau$ are chosen on validation either (a) by maximising recall under a false-alarm budget, or (b) by minimising FA/h subject to a target recall; selected $\tau$ are then frozen for testing. For clarity, FA/h can also be computed directly from test-set coverage (false positives per hour of labelled windows); unless otherwise noted, we report the scenario-translated FA/h using $(P,R,\lambda)$.

\subsection{Reference Model}
\label{subsec:ref-model}

\begin{figure}[!htbp]
  \centering
  \includegraphics[width=\linewidth,clip,trim=0 15mm 0 0]{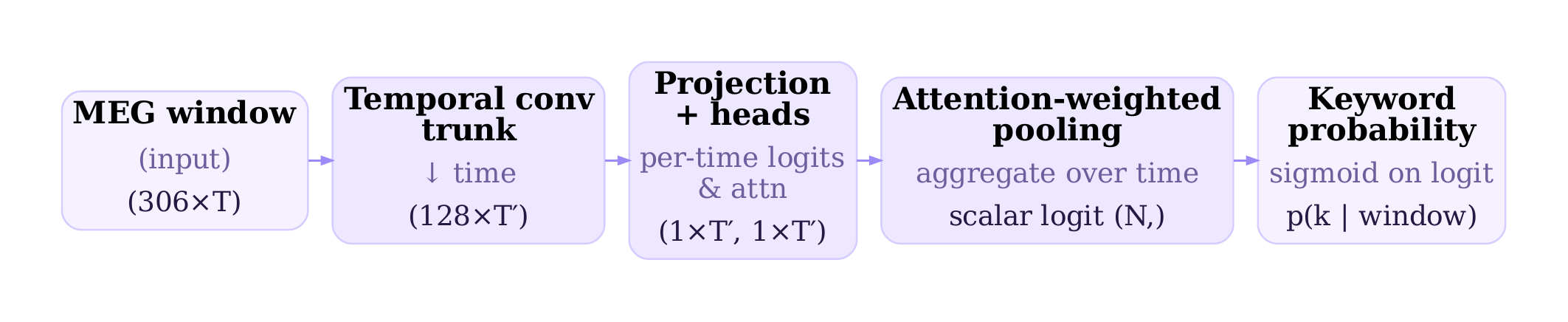}
  \caption{Reference model overview. A 306\,$\times$\,\textit{T} MEG window passes through a temporal convolutional trunk (with time downsampling) to produce a 128\,$\times$\,\textit{T'} representation. A projection stage feeds two temporal heads that emit per-time logits and attention scores. An attention-weighted pooling aggregates over time to a scalar logit, which is mapped to the keyword probability via a sigmoid.}
  \label{fig:ref-model}
\end{figure}

\vspace{-0.5\baselineskip}

\FloatBarrier

Our reference system ingests 306-channel MEG windows of length \textit{T} and processes them with a compact temporal convolutional trunk that includes a residual block and a time-downsampling layer, yielding a 
$128 \times \textit{T'}$ representation (temporal CNNs are strong sequence/biosignal decoders; residual connections stabilize deeper stacks and enlarge receptive fields efficiently \citep{bai2018tcn,schirrmeister2017deep,lawhern2018eegnet,he2016resnet}). A projection stage produces a 512-channel sequence, from which two $1 \times 1$ temporal heads compute (i) per-time logits and (ii) attention scores normalised along time. The final output is a scalar logit obtained by attention-weighted summation of the per-time logits (a learned MIL-style pooling well-suited to brief events within longer windows \citep{ilse2018attentionmil,kong2020panns,mcfee2018adaptivepooling}; in MEG, this lets the model emphasise time-locked acoustic/lexical responses such as M100/N400 components \citep{Gage1998_M100,Halgren2002_N400likeMEG,hari2012meg}). Training uses focal loss with a small pairwise ranking term: focal down-weights abundant easy negatives and focuses gradient on rare, hard positives under extreme imbalance \citep{lin2017focal}, while the pairwise (logistic) ranking aux loss encourages correct ordering of positives above negatives, supporting PR/Average-Precision-aligned selection \citep{burges2010ranknet,yue2007ap,davis2006prroc,saito2015prroc}. Batches are class-balanced by oversampling positives, and we apply light temporal jitter and additive noise (both standard, effective regularizers for EEG/MEG time-series \citep{buda2018imbalance,lashgari2020eegda,he2021eegda,rommel2022eegda}). We optimise with AdamW \citep{loshchilov2019adamw} and select checkpoints by validation AUPRC (preferred under heavy class imbalance \citep{saito2015prroc,davis2006prroc}).
\section{Results}
Where possible, all experiments use the standard train/validation/test splits provided by the \texttt{pnpl} dataset (using the logic described in Section \ref{subsec:ref-model}) and are fully reproducible (see Appendix \ref{appendix:code}). Unless noted, values are seed-averages over three runs. Error bars are standard errors across seeds. For Table \ref{tab:model-performance} we report standard errors approximated from 95\% bootstrap CIs (4,000 resamples).

\subsection{Model Performance}
\noindent We first establish that the dataset carries usable signal for keyword detection by evaluating a single model on the held-out test set (\(n=4660\), positives \(=24\); base rate \(=0.00515\)). Given the absence of prior publicly reproducible MEG keyword spotting benchmarks, we evaluate against permutation-derived random baselines to demonstrate the presence of meaningful signal rather than competitive performance. While overall performance is modest, threshold-free metrics indicate clear signal (AUPRC $\approx 13.4 \times$ the permutation baseline; AUROC $\approx 0.80$). Full results are provided in Table~\ref{tab:model-performance}.
\noindent Beyond threshold-free metrics, we include an operational snapshot. At a target recall of \(\sim\!0.10\), the scenario-translated FA/h for the assistive case ($\lambda=2$/h) is \(\sim\!2.19\) (SE \(\approx\!1.63\)), corresponding to \(\sim\!13\) alerts per correct detection. For reference, directly counting false positives per hour under the labelled test coverage yields \(\sim\!16.3\,\mathrm{FA/h}\) (seed-avg; SE \(\approx\!12.1\)). Under FA/h budgets (scenario scale, $\lambda=2$/h), the model achieves recall \(\sim\!0.14\) at 2.0 FA/h and \(\sim\!0.08\) at 0.5 FA/h (seed-averaged). These numbers contextualise the ranking metrics and set a baseline for future improvements. The right panel of Fig.~\ref{fig:keyword-length-uplift} shows the mean recall--FA/h operating curve with per-seed traces. For this snapshot, operating points are chosen on the test PR curves for presentation; in deployment we would select thresholds on validation and freeze them before testing.

\begin{table}[H]
\centering
\begin{tabular}{lcccc}
\hline
\textbf{Metric} & \textbf{Baseline} & \textbf{Model} (\(\pm\) SE) & \textbf{\% improvement} & \textbf{p-value} \\
\hline
F1 & 0.010 & \textbf{0.107} \( \pm \) 0.038 & +970\% & $\approx 1.00\times 10^{-5}$ \\
F1-Macro & 0.431 & \textbf{0.542} \( \pm \) 0.028 & +25.8\% & $\approx 1.00\times 10^{-5}$ \\
Accuracy & \textbf{0.995} & 0.955 \( \pm \) 0.033 & -4.0\% & n.s. \\
MCC  & 0.000 & \textbf{0.119} \( \pm \) 0.027 & n/a & $\approx 1.00\times 10^{-5}$ \\
AUROC                & 0.500 & \textbf{0.804} \( \pm \) 0.017 & +60.8\% & $\approx 2.00\times 10^{-5}$ \\
AUPRC                & 0.007 & \textbf{0.094} \( \pm \) 0.032 & +1243\% & $\approx 2.00\times 10^{-5}$ \\
\hline
\end{tabular}
\caption{Performance compared to random baselines derived from permutation nulls. Thresholded metrics use threshold \(\tau=0.5\). Standard errors are approximated from the 95\% bootstrap CIs via normality (\(\mathrm{SE} \approx ( \mathrm{CI}_{\mathrm{hi}} - \mathrm{CI}_{\mathrm{lo}} ) / 3.92\)).}
\label{tab:model-performance}
\end{table}


\subsection{Keyword Choice}
\label{sec:keyword-choice}
\begin{figure}[!htbp]
\centering
\includegraphics[width=0.85\textwidth]{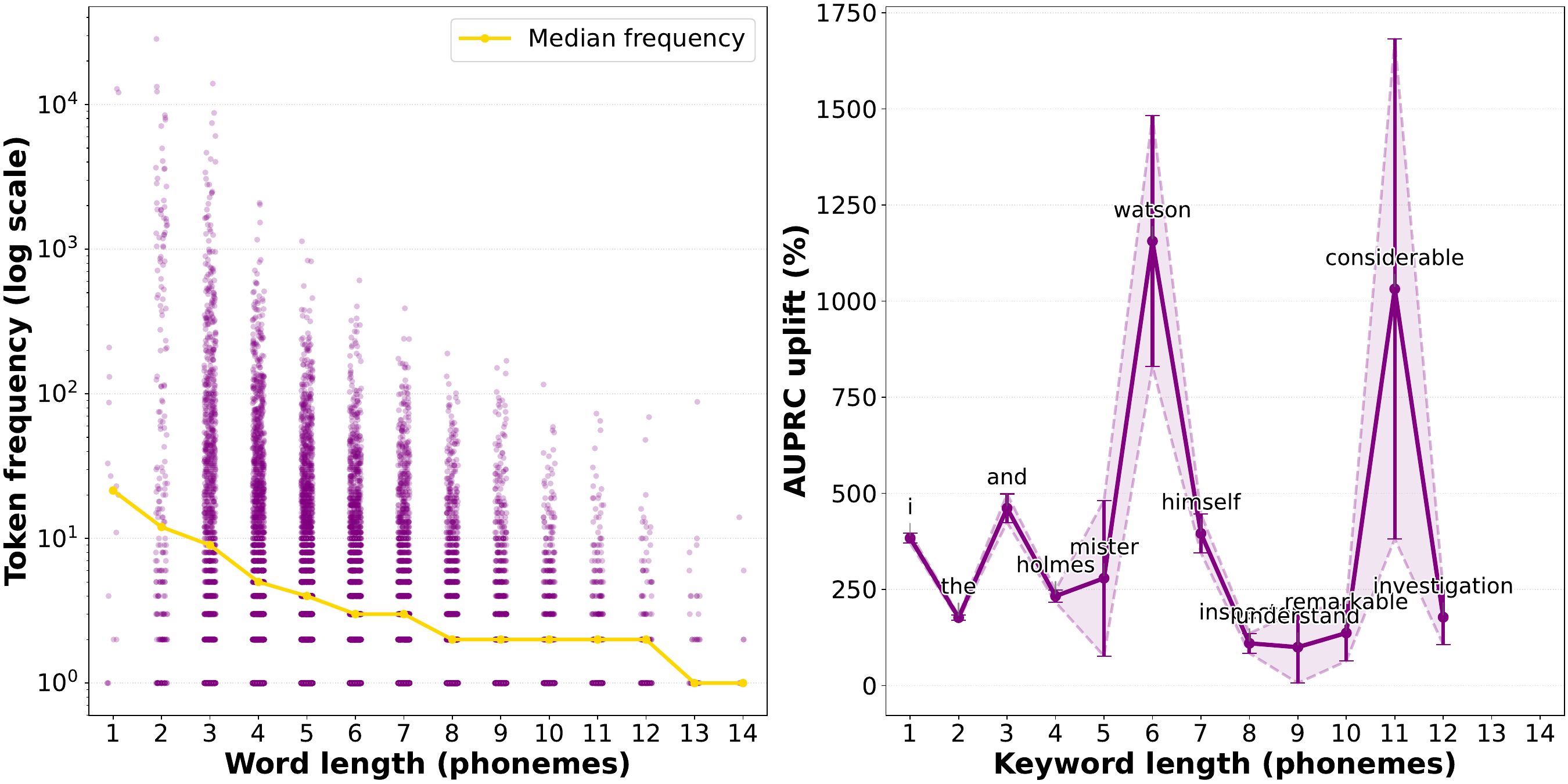}
\caption{Keyword choice trade-offs. (Left) Relationship between word length in phonemes and token frequency across the full LibriBrain corpus (points are unique words; y-axis is log-scaled for readability; the line shows the median frequency per length). (Right) \%$\Delta$AUPRC over the base rate for the shortlisted keywords as a function of their phoneme count.}
\label{fig:keyword-choice}
\end{figure}

An important consideration in KWS is the choise of keyword(s). In LibriBrain, as in many real-world corpora, longer words are rarer: word length in phonemes is negatively correlated with token frequency (Spearman $r=-0.28$ with $\log$ frequency; $p=2.7\times 10^{-185}$; Fig.~\ref{fig:keyword-choice} left). This matters because length can reduce false alarms while frequency controls how many positives we can realistically train on. To navigate this trade-off, we selected the most frequent word at each phoneme length and measured \%$\Delta$AUPRC over the empirical base rate. The length-\%$\Delta$AUPRC relation is non-monotonic (Fig.~\ref{fig:keyword-choice} right): among a 12-item shortlist spanning 1-12 phonemes, the 5-phoneme \emph{watson} yields the largest \%$\Delta$AUPRC, whereas several longer items (e.g., 9-12 phonemes) underperform despite greater duration.

\begin{figure}[!htbp]
\centering
\begin{minipage}{0.43\textwidth}
\centering
\includegraphics[width=\linewidth]{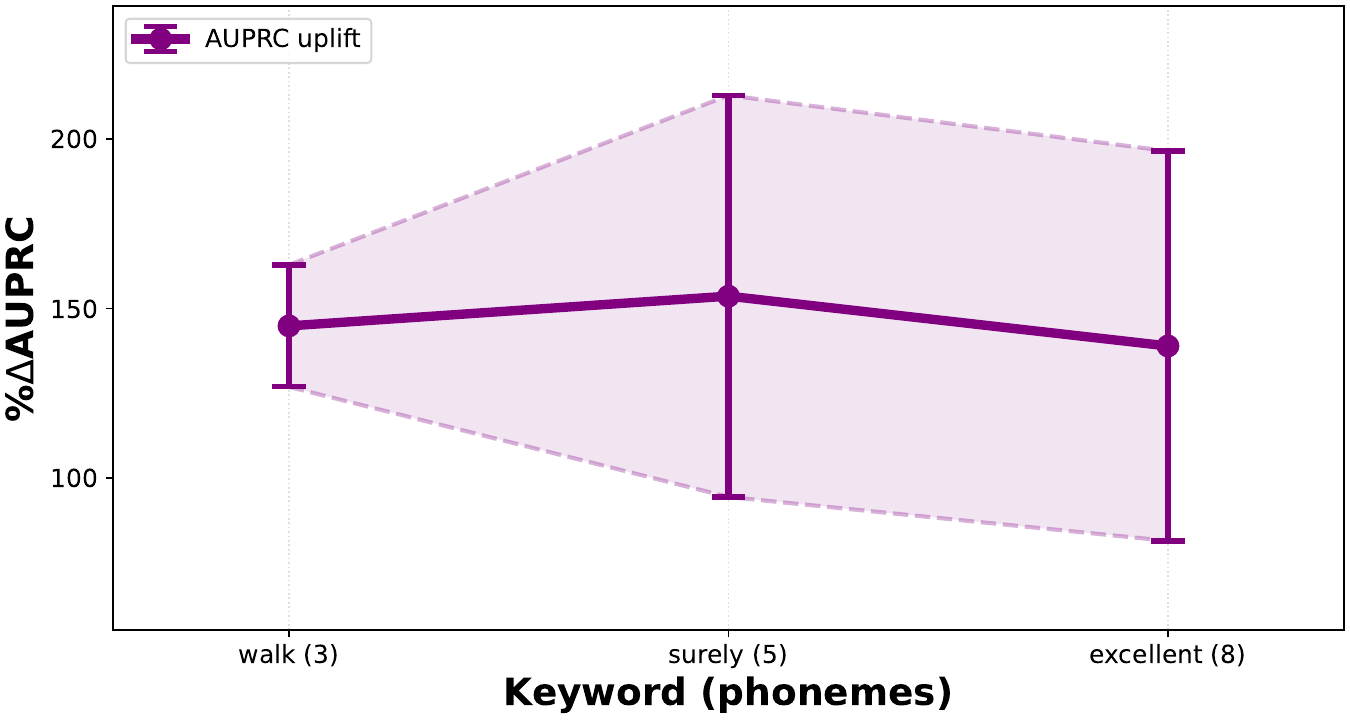}
\end{minipage}\hfill
\begin{minipage}{0.43\textwidth}
\centering
\includegraphics[width=\linewidth]{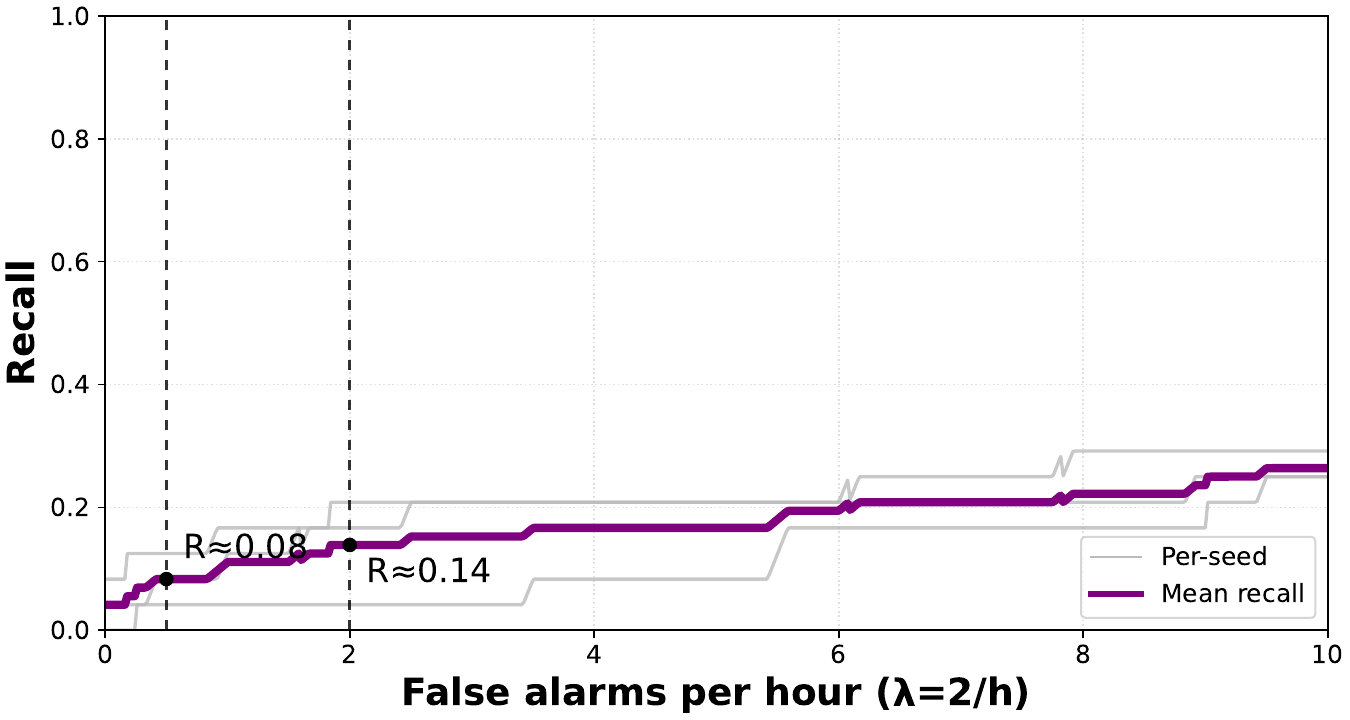}
\end{minipage}
\caption{Left: \%$\Delta$AUPRC over the empirical base rate for three similarly frequent keywords with 3, 5, and 8 phonemes, showing no significant difference. Right: Recall--FA/h operating curve (assistive, $\lambda=2$/h) with budget markers at 0.5 and 2.0 FA/h.}
\label{fig:keyword-length-uplift}
\end{figure}

A controlled comparison across three similarly frequent keywords of different lengths (\emph{walk}, \emph{surely}, \emph{excellent}; 3/5/8 phonemes) shows no detectable difference in \%$\Delta$AUPRC within our precision (overlapping SEMs; Fig.~\ref{fig:keyword-length-uplift}). This indicates that, once frequency is matched, mere length is not the primary driver of detectability in MEG KWS. The pattern is consistent with established constraints on neural speech processing and KWS: benefits accrue less from duration per se and more from properties that improve time-locking and reduce lexical competition—salient acoustic onsets and early stress (stronger M100/M200),  an early uniqueness point (UP)—i.e., the keyword becomes lexically unique after only a few initial phonemes (a small UP index relative to its length)—a sparse phonological neighborhood, and moderate frequency with lower contextual predictability \citep{Gage1998_M100,Leminen2011_UP_MEGEEG,VitevitchLuce1999_JML,Halgren2002_N400likeMEG,Chen2014_SmallFootprintKWS}. Empirically, \emph{watson} may profit from prosodic prominence in narrative speech and an early uniqueness point, outweighing any gains attributable to length alone; \textit{watson} may also benefit from attentional saliency, being a word that the subject consistently paid attention to. 

\FloatBarrier

\subsection{Data Scaling}
\noindent To understand how keyword detection performance scales with available training data, we systematically varied the fraction of the 52-hour corpus used for training while keeping the validation and test sets fixed. For these scaling runs we used 0\,s pre-onset and +0.25\,s post-onset windows (per-instance window length of 1.05\,s). Because training uses many overlapping windows around labelled events, the total windowed duration processed exceeds the 52\,h of unique recordings: at 10\% this corresponds to \(\approx\)14\,h of windowed data (\(\approx\)5.2\,h unique), and at 100\% to \(\approx\)143\,h of windowed data.

\begin{figure}[!htbp]
\centering
\begin{minipage}{0.49\textwidth}
\centering
\includegraphics[width=\linewidth]{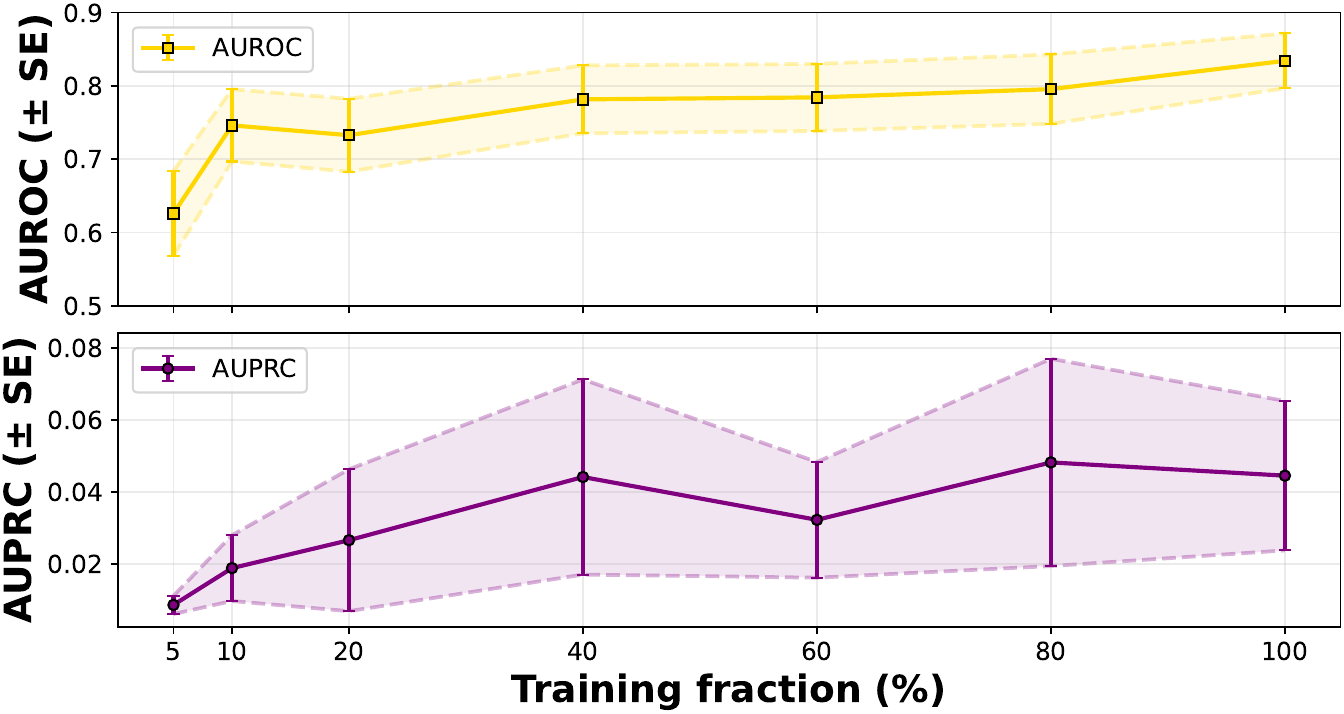}
\end{minipage}\hfill
\begin{minipage}{0.49\textwidth}
\centering
\includegraphics[width=\linewidth]{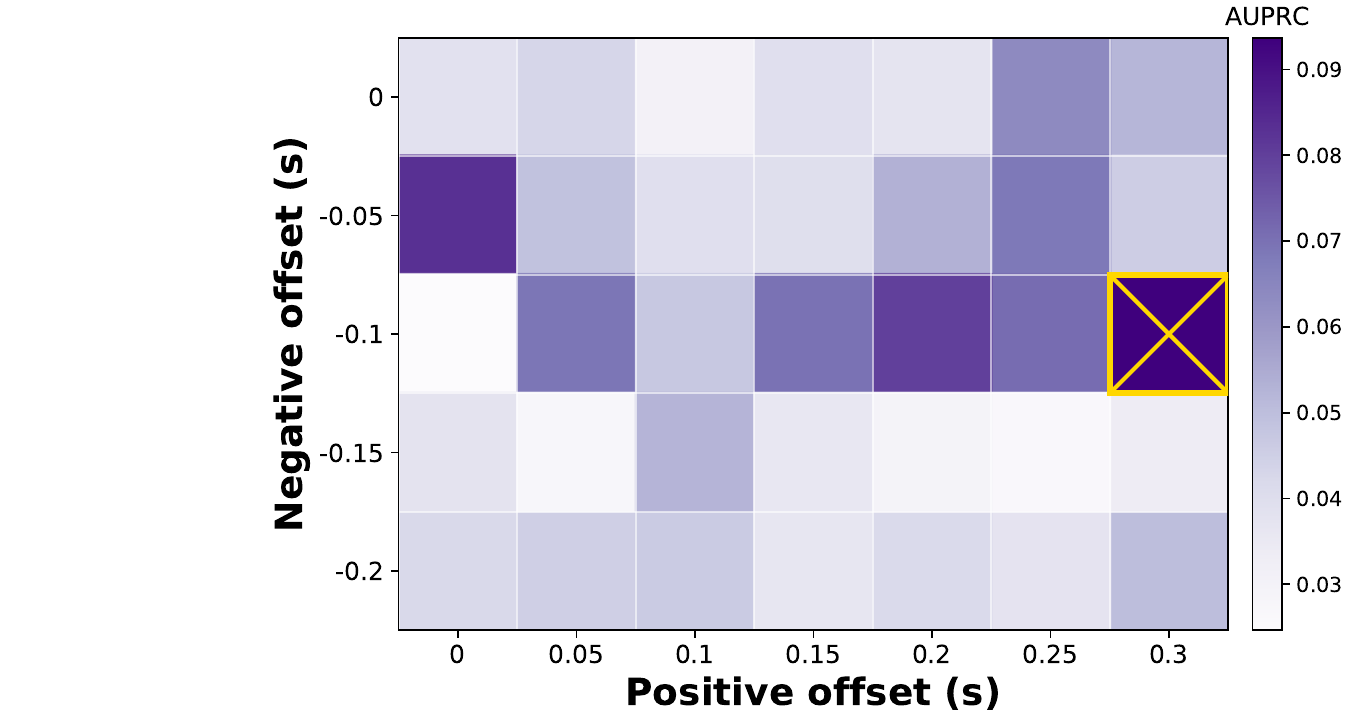}
\end{minipage}
\caption{Left: Keyword spotting scales with the amount of training data; AUPRC improves roughly log-linearly as a larger fraction of the 52-hour corpus is used. Right: Effect of temporal offsets around the keyword onset. The X marks where AUPRC peaks with a modest pre-onset context (\(\sim\!0.1\)s) and a slightly longer post-onset window (\(\sim\!0.3\)s).}
\label{fig:data-scaling}
\label{fig:buffer-heatmap}
\end{figure}

As shown in Figure~\ref{fig:data-scaling}, AUPRC improves approximately log-linearly as we increase the training fraction from 10\% to 100\%, consistent with established within-subject scaling laws in neural decoding \citep{dascoli2024words, sato2024scaling}. Notably, even with just 10\% of the training data (\(\approx\)14\,h windowed; \(\approx\)5.2\,h unique), the model achieves meaningful performance above chance, suggesting that keyword detection remains feasible even in scenarios with limited recording time. Permutation tests confirm that AUPRC is not above chance at 5\% (\(p=0.108\)), but is already significant at 10\% (\(p=0.0156\); one-sided, 10{,}000 draws), and remains strongly significant thereafter (20\% \(p=6.0\times 10^{-4}\); 40--100\% \(p\le 2\times 10^{-4}\)).

\subsection{Sample Length}

\noindent Adding pre- and post-onset offsets modestly improves detection. Averaged over all non-zero offsets, AUPRC increases by \(\sim\!25\%\) relative to the 0/0 baseline (absolute +0.0097). This improvement is statistically significant (paired per-seed mean +0.0099 AUPRC, SE 0.0028; 95\% CI [0.0045, 0.0154]; one-sided \(p<10^{-3}\)). We observe the best AUPRC near (neg=0.1s, pos=0.3s), but no clear monotonic trend across offsets; performance is relatively flat in a small neighbourhood around this setting and declines for very short or overly long windows, suggesting a bias--variance trade-off between providing sufficient context and diluting signal with unrelated activity.

\FloatBarrier



\section{Conclusion}
We introduce a reproducible MEG keyword-spotting task on LibriBrain, demonstrate meaningful signal, and release task specifications, a modified \texttt{pnpl} library, baseline model, and tutorial materials.

\subsection{Practical Utility}
Despite the achieved improvement in metrics, performance is not yet sufficient for reliable hands-free use. In an assistive scenario ($\lambda{=}2\,\mathrm{h}^{-1}$), at recall $\approx\!0.10$ the system yields $\approx\!2.2$ false alarms per hour (about 13 alerts per correct detection). Priorities for future work thus include: (i) stronger ranking, (ii) calibration and principled threshold selection, and (iii) deployment strategies that suppress false alarms (multi-confirmation, small ensembles, cascaded detectors with context-aware priors).

\subsection{Limitations \& Future Work}
\paragraph{Limitations.} This study reports results for a single participant on a single corpus. Generalisability across participants remains an open question for Neural Keyword Spotting, though generalization is becoming less of a problem in decoding than it was \citep{csaky2023group, defossez2023nmi, dascoli2024words, jayalath2025icml, jayalath2025unlocking}. Validation and test sessions were selected to maximise positives, stabilising metrics at the expense of a mild base-rate bias. Test sets contain few positives, so thresholded metrics carry substantial uncertainty. We evaluate a compact model only, without exploring richer encoders, self-supervised pretraining, streaming inference, or multi-keyword training. Finally, we use event-referenced windows; continuous-stream detection with latency and explicit false-alarm accounting remains open.

\paragraph{Future work.} Building on the success of competitions like the 2025 LibriBrain Competition \citep{landau2025competition} or Brain-to-text '25 \citep{card2025brain2text}, we will release a leaderboard system for the keyword detection task as part of a larger-scale competition later this year. We also plan to extend this work to additional datasets, including inner speech and multi-subject recordings. Further analysis, such as phoneme-informed keyword selection and a deeper examination of temporal offsets, should clarify where gains are available building on the exploratory results presented here.

\begin{ack}
We would like to thank to the organizers and anonymous reviewers for their efforts and constructive feedback. We also acknowledge the use of the University of Oxford Advanced Research Computing (ARC) facility (\url{http: //dx.doi.org/10.5281/zenodo.22558}) and the NVIDIA Corporation for contributing additional GPUs. PNPL is supported by the MRC (MR/X00757X/1), Royal Society (RG\textbackslash
R1\textbackslash
241267), NSF (2314493), NFRF (NFRFT-2022-00241), and SSHRC (895-2023-1022).

\end{ack}

\bibliographystyle{plainnat}
\bibliography{references}

\begin{thebibliography}{62}
\providecommand{\natexlab}[1]{#1}
\providecommand{\url}[1]{\texttt{#1}}
\expandafter\ifx\csname urlstyle\endcsname\relax
  \providecommand{\doi}[1]{doi: #1}\else
  \providecommand{\doi}{doi: \begingroup \urlstyle{rm}\Url}\fi

\bibitem[Anthropic(2025)]{anthropic2025claude37}
Anthropic.
\newblock Claude 3.7 sonnet and claude code.
\newblock \url{https://www.anthropic.com/news/claude-3-7-sonnet}, 2025.
\newblock Accessed: 2025-04-22.

\bibitem[{Apple Siri Team}(2017)]{apple2017heysiri}
{Apple Siri Team}.
\newblock Hey {S}iri: An on-device {DNN}-powered voice trigger for {A}pple's personal assistant.
\newblock Apple Machine Learning Journal, 2017.
\newblock URL \url{https://machinelearning.apple.com/research/hey-siri}.

\bibitem[Ar{\'\i}k et~al.(2017)Ar{\'\i}k, Kliegl, Child, Hestness, Gibiansky, Fougner, Prenger, and Coates]{arik2017crnn}
Sercan~{\"O}. Ar{\'\i}k, Markus Kliegl, Rewon Child, Joel Hestness, Andrew Gibiansky, Chris Fougner, Ryan Prenger, and Adam Coates.
\newblock Convolutional recurrent neural networks for small-footprint keyword spotting.
\newblock In \emph{Interspeech}, pages 1606--1610, 2017.
\newblock \doi{10.21437/Interspeech.2017-1737}.

\bibitem[Bai et~al.(2018)Bai, Kolter, and Koltun]{bai2018tcn}
Shaojie Bai, J.~Zico Kolter, and Vladlen Koltun.
\newblock An empirical evaluation of generic convolutional and recurrent networks for sequence modeling.
\newblock \emph{arXiv preprint arXiv:1803.01271}, 2018.

\bibitem[Berg et~al.(2021)Berg, O'Connor, and Tairum~Cruz]{howard2021kwt}
Axel Berg, Mark O'Connor, and Miguel Tairum~Cruz.
\newblock Keyword transformer: Auditory attention for small-footprint keyword spotting.
\newblock \emph{arXiv preprint arXiv:2104.00769}, 2021.

\bibitem[Buda et~al.(2018)Buda, Maki, and Mazurowski]{buda2018imbalance}
Mateusz Buda, Atsuto Maki, and Maciej~A. Mazurowski.
\newblock A systematic study of the class imbalance problem in convolutional neural networks.
\newblock \emph{Neural Networks}, 106:\penalty0 249--259, 2018.
\newblock \doi{10.1016/j.neunet.2018.07.011}.

\bibitem[Burges(2010)]{burges2010ranknet}
Christopher J.~C. Burges.
\newblock From {RankNet} to {LambdaRank} to {LambdaMART}: An overview.
\newblock Technical Report MSR-TR-2010-82, Microsoft Research, 2010.

\bibitem[Card et~al.(2025)Card, Wairagkar, Iacobacci, Hou, Singer-Clark, Willett, Kunz, Fan, Nia, Deo, Srinivasan, Choi, Glasser, Hochberg, Henderson, Shahlaie, Stavisky, and Brandman]{card2025brain2text}
Nicholas Card, Maitreyee Wairagkar, Carrina Iacobacci, Xianda Hou, Tyler Singer-Clark, Francis~R. Willett, Erin~M. Kunz, Chaofei Fan, Maryam~Vahdati Nia, Darrel~R. Deo, Aparna Srinivasan, Eun~Young Choi, Matthew~F. Glasser, Leigh~R. Hochberg, Jaimie~M. Henderson, Kiarash Shahlaie, Sergey~D. Stavisky, and David~M. Brandman.
\newblock Brain-to-text '25.
\newblock Kaggle, 2025.
\newblock URL \url{https://kaggle.com/competitions/brain-to-text-25}.

\bibitem[Chen et~al.(2014)Chen, Parada, and Heigold]{Chen2014_SmallFootprintKWS}
Guoguo Chen, Carolina Parada, and Georg Heigold.
\newblock Small-footprint keyword spotting using deep neural networks.
\newblock In \emph{2014 IEEE International Conference on Acoustics, Speech and Signal Processing (ICASSP)}, pages 4087--4091, 2014.
\newblock \doi{10.1109/ICASSP.2014.6854370}.

\bibitem[Csaky et~al.(2023)Csaky, van Es, Parker~Jones, and Woolrich]{csaky2023group}
Richard Csaky, Mats W.~J. van Es, Oiwi Parker~Jones, and Mark Woolrich.
\newblock Group-level brain decoding with deep learning.
\newblock \emph{Human Brain Mapping}, 44:\penalty0 6105--6119, 2023.
\newblock \doi{10.1002/hbm.26500}.
\newblock URL \url{https://doi.org/10.1002/hbm.26500}.

\bibitem[d'Ascoli et~al.(2024)d'Ascoli, Bel, Rapin, Banville, Benchetrit, Pallier, and King]{dascoli2024words}
St{\'e}phane d'Ascoli, Corentin Bel, J{\'e}r{\'e}my Rapin, Hubert Banville, Yohann Benchetrit, Christophe Pallier, and Jean-R{\'e}mi King.
\newblock Decoding individual words from non-invasive brain recordings across 723 participants.
\newblock \emph{arXiv preprint arXiv:2412.17829}, 2024.

\bibitem[Dash et~al.(2020)Dash, Ferrari, Dutta, and Wang]{Dash2020NeuroVAD}
Debadatta Dash, Paul Ferrari, Satwik Dutta, and Jun Wang.
\newblock Neurovad: Real-time voice activity detection from non-invasive neuromagnetic signals.
\newblock \emph{Sensors}, 20\penalty0 (8):\penalty0 2248, 2020.
\newblock \doi{10.3390/s20082248}.

\bibitem[Davis and Goadrich(2006)]{davis2006prroc}
Jesse Davis and Mark Goadrich.
\newblock The relationship between precision-recall and {ROC} curves.
\newblock In \emph{Proceedings of the 23rd International Conference on Machine Learning (ICML)}, pages 233--240, 2006.
\newblock \doi{10.1145/1143844.1143874}.

\bibitem[Desai et~al.(2021)Desai, Holder, Villarreal, Clark, Hoang, and Hamilton]{desai2021generalizable}
Maansi Desai, Jade Holder, Cassandra Villarreal, Nat Clark, Brittany Hoang, and Liberty~S Hamilton.
\newblock Generalizable {EEG} encoding models with naturalistic audiovisual stimuli.
\newblock \emph{Journal of Neuroscience}, 41\penalty0 (43):\penalty0 8946--8962, 2021.

\bibitem[Duan et~al.(2023)Duan, Zhou, Wang, Wang, and Lin]{duan2023dewave}
Yiqun Duan, Jinzhao Zhou, Zhen Wang, Yu-Kai Wang, and Chin-Teng Lin.
\newblock {DeWave}: Discrete {EEG} waves encoding for brain dynamics to text translation.
\newblock \emph{arXiv preprint arXiv:2309.14030}, 2023.

\bibitem[Défossez et~al.(2023)Défossez, Caucheteux, Rapin, Kabeli, and King]{defossez2023nmi}
Alexandre Défossez, Charlotte Caucheteux, Jérémy Rapin, Ori Kabeli, and Jean-Rémi King.
\newblock Decoding speech perception from non-invasive brain activity with self-supervised learning.
\newblock \emph{Nature Machine Intelligence}, 5\penalty0 (11):\penalty0 1262--1276, 2023.
\newblock \doi{10.1038/s42256-023-00732-3}.

\bibitem[Gage et~al.(1998)Gage, Poeppel, Roberts, and Hickok]{Gage1998_M100}
Nicole Gage, David Poeppel, Timothy P.~L. Roberts, and Gregory Hickok.
\newblock Auditory evoked m100 reflects onset acoustics of speech sounds.
\newblock \emph{Brain Research}, 814:\penalty0 236--239, 1998.
\newblock \doi{10.1016/S0006-8993(98)01058-0}.

\bibitem[Gwilliams et~al.(2022)Gwilliams, King, Marantz, and Poeppel]{gwilliams2022phonseq}
Laura Gwilliams, Jean-R{\'e}mi King, Alec Marantz, and David Poeppel.
\newblock Neural dynamics of phoneme sequences reveal position-invariant code for content and order.
\newblock \emph{Nature Communications}, 13\penalty0 (1):\penalty0 6606, 2022.
\newblock \doi{10.1038/s41467-022-34326-1}.

\bibitem[Gwilliams et~al.(2023)Gwilliams, Flick, Marantz, Pylkkänen, Poeppel, and King]{gwilliams2023megmasc}
Laura Gwilliams, Graham Flick, Alec Marantz, Liina Pylkkänen, David Poeppel, and Jean-Rémi King.
\newblock {MEG-MASC}: A magnetoencephalography dataset for naturalistic language comprehension.
\newblock \emph{Scientific Data}, 10\penalty0 (1):\penalty0 1--16, 2023.
\newblock \doi{10.1038/s41597-023-02170-x}.

\bibitem[Halgren et~al.(2002)Halgren, Dhond, Christensen, Van~Petten, Marinkovic, Lewine, and Dale]{Halgren2002_N400likeMEG}
Eric Halgren, Rupali Dhond, Niels Christensen, Cyma Van~Petten, Ksenija Marinkovic, Jeffrey Lewine, and Anders~M. Dale.
\newblock N400-like magnetoencephalography responses modulated by semantic context, word frequency, and lexical class in sentences.
\newblock \emph{NeuroImage}, 17\penalty0 (3):\penalty0 1101--1116, 2002.
\newblock \doi{10.1006/nimg.2002.1268}.

\bibitem[Hari and Salmelin(2012)]{hari2012meg}
Riitta Hari and Riitta Salmelin.
\newblock Magnetoencephalography: From {SQUIDs} to neuroscience.
\newblock \emph{NeuroImage}, 61\penalty0 (2):\penalty0 386--396, 2012.
\newblock \doi{10.1016/j.neuroimage.2011.11.074}.

\bibitem[He et~al.(2021)He, Liu, Zhu, and Du]{he2021eegda}
Chao He, Jialu Liu, Yuesheng Zhu, and Wencai Du.
\newblock Data augmentation for deep neural networks model in {EEG} classification task: A review.
\newblock \emph{Frontiers in Human Neuroscience}, 15:\penalty0 765525, 2021.
\newblock \doi{10.3389/fnhum.2021.765525}.

\bibitem[He et~al.(2016)He, Zhang, Ren, and Sun]{he2016resnet}
Kaiming He, Xiangyu Zhang, Shaoqing Ren, and Jian Sun.
\newblock Deep residual learning for image recognition.
\newblock In \emph{Proceedings of the IEEE Conference on Computer Vision and Pattern Recognition (CVPR)}, pages 770--778, 2016.
\newblock \doi{10.1109/CVPR.2016.90}.

\bibitem[Ilse et~al.(2018)Ilse, Tomczak, and Welling]{ilse2018attentionmil}
Maximilian Ilse, Jakub Tomczak, and Max Welling.
\newblock Attention-based deep multiple instance learning.
\newblock In Jennifer Dy and Andreas Krause, editors, \emph{Proceedings of the 35th International Conference on Machine Learning (ICML)}, volume~80 of \emph{Proceedings of Machine Learning Research}, pages 2127--2136. PMLR, Jul 2018.
\newblock URL \url{https://proceedings.mlr.press/v80/ilse18a.html}.

\bibitem[Jayalath et~al.(2025{\natexlab{a}})Jayalath, Landau, and Parker~Jones]{jayalath2025unlocking}
Dulhan Jayalath, Gilad Landau, and Oiwi Parker~Jones.
\newblock Unlocking non-invasive brain-to-text.
\newblock \emph{arXiv preprint arXiv:2505.13446}, 2025{\natexlab{a}}.

\bibitem[Jayalath et~al.(2025{\natexlab{b}})Jayalath, Landau, Shillingford, Woolrich, and Parker~Jones]{jayalath2025icml}
Dulhan Jayalath, Gilad Landau, Brendan Shillingford, Mark Woolrich, and Oiwi Parker~Jones.
\newblock {The Brain's Bitter Lesson}: Scaling speech decoding with self-supervised learning.
\newblock \emph{International Conference on Machine Learning (ICML)}, 2025{\natexlab{b}}.

\bibitem[Jo et~al.(2024)Jo, Yang, Han, Duan, Xiong, and Lee]{jo2024eegtotext}
Hyejeong Jo, Yiqian Yang, Juhyeok Han, Yiqun Duan, Hui Xiong, and Won~Hee Lee.
\newblock Are {EEG}-to-text models working?
\newblock \emph{arXiv preprint}, 2024.
\newblock URL \url{https://arxiv.org/abs/2405.06459}.

\bibitem[Kong et~al.(2020)Kong, Cao, Iqbal, Wang, Wang, and Plumbley]{kong2020panns}
Qiuqiang Kong, Yin Cao, Turab Iqbal, Yuxuan Wang, Wenwu Wang, and Mark~D. Plumbley.
\newblock {PANNs}: Large-scale pretrained audio neural networks for audio pattern recognition.
\newblock \emph{IEEE/ACM Transactions on Audio, Speech, and Language Processing}, 28:\penalty0 2880--2894, 2020.
\newblock \doi{10.1109/TASLP.2020.3030497}.

\bibitem[Kunz et~al.(2025)Kunz, Krasa, Kamdar, Avansino, Hahn, Yoon, Singh, Nason-Tomaszewski, Card, Jude, Jacques, Bechefsky, Iacobacci, Hochberg, Rubin, Williams, Brandman, Stavisky, AuYong, Pandarinath, Druckmann, Henderson, and Willett]{kunz2025inner}
Erin~M. Kunz, Benyamin~Abramovich Krasa, Foram Kamdar, Donald~T. Avansino, Nick Hahn, Seonghyun Yoon, Akansha Singh, Samuel~R. Nason-Tomaszewski, Nicholas~S. Card, Justin~J. Jude, Brandon~G. Jacques, Payton~H. Bechefsky, Carrina Iacobacci, Leigh~R. Hochberg, Daniel~B. Rubin, Ziv~M. Williams, David~M. Brandman, Sergey~D. Stavisky, Nicholas AuYong, Chethan Pandarinath, Shaul Druckmann, Jaimie~M. Henderson, and Francis~R. Willett.
\newblock Inner speech in motor cortex and implications for speech neuroprostheses.
\newblock \emph{Cell}, 188\penalty0 (17):\penalty0 4658--4673.e17, 2025.
\newblock \doi{10.1016/j.cell.2025.06.015}.

\bibitem[Landau et~al.(2025)Landau, Özdogan, Elvers, Mantegna, Somaiya, Jayalath, Kurth, Kwon, Shillingford, Farquhar, Jiang, Jerbi, Abdelhedi, Mantilla~Ramos, Gulcehre, Woolrich, Voets, and Parker~Jones]{landau2025competition}
Gilad Landau, Miran Özdogan, Gereon Elvers, Francesco Mantegna, Pratik Somaiya, Dulhan Jayalath, Luisa Kurth, Teyun Kwon, Brendan Shillingford, Greg Farquhar, Minqi Jiang, Karim Jerbi, Hamza Abdelhedi, Yorguin Mantilla~Ramos, Caglar Gulcehre, Mark Woolrich, Natalie Voets, and Oiwi Parker~Jones.
\newblock The 2025 {PNPL} competition: Speech detection and phoneme classification in the {LibriBrain} dataset.
\newblock \emph{NeurIPS, Competition Track}, 2025.
\newblock \url{https://arxiv.org/abs/2506.10165}.

\bibitem[Lashgari et~al.(2020)Lashgari, Liang, and Maoz]{lashgari2020eegda}
Elnaz Lashgari, Dehua Liang, and Uri Maoz.
\newblock Data augmentation for deep-learning-based electroencephalography.
\newblock \emph{Journal of Neuroscience Methods}, 346:\penalty0 108885, 2020.
\newblock \doi{10.1016/j.jneumeth.2020.108885}.

\bibitem[Lawhern et~al.(2018)Lawhern, Solon, Waytowich, Gordon, Hung, and Lance]{lawhern2018eegnet}
Vernon~J. Lawhern, Amelia~J. Solon, Nicholas~R. Waytowich, Stephen~M. Gordon, Chou~P. Hung, and Brent~J. Lance.
\newblock {EEGNet}: A compact convolutional neural network for {EEG}-based brain–computer interfaces.
\newblock \emph{Journal of Neural Engineering}, 15\penalty0 (5):\penalty0 056013, 2018.
\newblock \doi{10.1088/1741-2552/aace8c}.

\bibitem[Leminen et~al.(2011)Leminen, Leminen, Kujala, and Shtyrov]{Leminen2011_UP_MEGEEG}
Aleksi Leminen, Mikko Leminen, Teija Kujala, and Yury Shtyrov.
\newblock A combined {EEG} and {MEG} study of spoken word recognition time-locked to the uniqueness point.
\newblock \emph{Frontiers in Human Neuroscience}, 5:\penalty0 66, 2011.
\newblock \doi{10.3389/fnhum.2011.00066}.

\bibitem[Lin et~al.(2017)Lin, Goyal, Girshick, He, and Doll{\'a}r]{lin2017focal}
Tsung-Yi Lin, Priya Goyal, Ross Girshick, Kaiming He, and Piotr Doll{\'a}r.
\newblock Focal loss for dense object detection.
\newblock In \emph{Proceedings of the IEEE International Conference on Computer Vision (ICCV)}, pages 2980--2988, 2017.
\newblock \doi{10.1109/ICCV.2017.324}.

\bibitem[Loshchilov and Hutter(2019)]{loshchilov2019adamw}
Ilya Loshchilov and Frank Hutter.
\newblock Decoupled weight decay regularization.
\newblock In \emph{International Conference on Learning Representations (ICLR)}, 2019.
\newblock URL \url{https://openreview.net/forum?id=Bkg6RiCqY7}.

\bibitem[Maidina et~al.(2020)]{matchboxnet2020}
Asmaa Maidina et~al.
\newblock Matchboxnet: 1d time-channel separable convolutional neural network architecture for speech commands recognition.
\newblock \emph{arXiv preprint arXiv:2004.08531}, 2020.

\bibitem[McFee et~al.(2018)McFee, Lostanlen, Salamon, Cartwright, and Bello]{mcfee2018adaptivepooling}
Brian McFee, Vincent Lostanlen, Justin Salamon, Mark Cartwright, and Juan~Pablo Bello.
\newblock Adaptive pooling operators for weakly labeled sound event detection.
\newblock In \emph{2018 IEEE International Conference on Acoustics, Speech and Signal Processing (ICASSP)}. IEEE, 2018.
\newblock \doi{10.1109/ICASSP.2018.8462220}.

\bibitem[Metzger et~al.(2023)Metzger, Littlejohn, Silva, Moses, Seaton, Wang, Dougherty, Liu, Wu, Berger, Zhuravleva, Tu-Chan, Ganguly, Anumanchipalli, and Chang]{metzger2023avatar}
Sean~L. Metzger, Kaylo~T. Littlejohn, Alexander~B. Silva, David~A. Moses, Margaret~P. Seaton, Ran Wang, Maximilian~E. Dougherty, Jessie~R. Liu, Peter Wu, Michael~A. Berger, Inga Zhuravleva, Adelyn Tu-Chan, Karunesh Ganguly, Gopala~K. Anumanchipalli, and Edward~F. Chang.
\newblock A high-performance neuroprosthesis for speech decoding and avatar control.
\newblock \emph{Nature}, 620\penalty0 (7976):\penalty0 1037--1046, 2023.
\newblock \doi{10.1038/s41586-023-06443-4}.

\bibitem[Milsap et~al.(2019)Milsap, Collard, Coogan, Rabbani, Wang, and Crone]{milsap2019keyword}
Griffin Milsap, Maxwell Collard, Christopher Coogan, Qinwan Rabbani, Yujing Wang, and Nathan~E. Crone.
\newblock Keyword spotting using human electrocorticographic recordings.
\newblock \emph{Frontiers in Neuroscience}, 13:\penalty0 60, 2019.
\newblock \doi{10.3389/fnins.2019.00060}.

\bibitem[Nastase et~al.(2022)Nastase, Liu, Hillman, Zadbood, Hasenfratz, Keshavarzian, Chen, Honey, Yeshurun, Regev, Nguyen, Chang, Baldassano, Lositsky, Simony, Chow, Leong, Brooks, Micciche, Choe, Goldstein, Vanderwal, Halchenko, Norman, and Hasson]{armeni2022narratives}
Samuel~A. Nastase, Yun-Fei Liu, Hanna Hillman, Asieh Zadbood, Liat Hasenfratz, Neggin Keshavarzian, Janice Chen, Christopher~J. Honey, Yaara Yeshurun, Mor Regev, Mai Nguyen, Claire H.~C. Chang, Christopher Baldassano, Olga Lositsky, Erez Simony, Michael~A. Chow, Yuan~Chang Leong, Paula~P. Brooks, Emily Micciche, Gina Choe, Ariel Goldstein, Tamara Vanderwal, Yaroslav~O. Halchenko, Kenneth~A. Norman, and Uri Hasson.
\newblock The narratives collection: Shared annotated datasets for naturalistic language processing in the neuroimaging of story listening.
\newblock \emph{Scientific Data}, 9\penalty0 (1):\penalty0 1--23, 2022.
\newblock \doi{10.1038/s41597-022-01735-6}.

\bibitem[Ochshorn and Hawkins(2015)]{gentle}
Robert~M. Ochshorn and Max Hawkins.
\newblock {Gentle}: A robust yet lenient forced aligner built on {Kaldi}, 2015.
\newblock URL \url{https://lowerquality.com/gentle/}.

\bibitem[\"Ozdogan et~al.(2025)\"Ozdogan, Landau, Elvers, Jayalath, Somaiya, Mantegna, Woolrich, and Jones]{ozdogan2025dataset}
Miran \"Ozdogan, Gilad Landau, Gereon Elvers, Dulhan Jayalath, Pratik Somaiya, Francesco Mantegna, Mark Woolrich, and Oiwi~Parker Jones.
\newblock {LibriBrain}: Over 50 hours of within-subject {MEG} to improve speech decoding methods at scale.
\newblock \emph{NeurIPS, Datasets \& Benchmarks Track}, 2025.
\newblock URL \url{https://arxiv.org/abs/2506.02098}.

\bibitem[Ridge and Parker~Jones(2024)]{ridge2024domainshift}
Jeremy Ridge and Oiwi Parker~Jones.
\newblock Resolving domain shift for representations of speech in non-invasive brain recordings.
\newblock \emph{arXiv preprint}, 2024.
\newblock URL \url{https://arxiv.org/abs/2410.19986}.

\bibitem[Rommel et~al.(2022)Rommel, Paillard, Moreau, and Gramfort]{rommel2022eegda}
C{\'e}dric Rommel, Joseph Paillard, Thomas Moreau, and Alexandre Gramfort.
\newblock Data augmentation for learning predictive models on {EEG}: a systematic comparison.
\newblock \emph{Journal of Neural Engineering}, 19\penalty0 (6), 2022.
\newblock \doi{10.1088/1741-2552/aca220}.

\bibitem[Russakovsky et~al.(2015)Russakovsky, Deng, Su, Krause, Satheesh, Ma, Huang, Karpathy, Khosla, Bernstein, Berg, and Fei-Fei]{russakovsky2015imagenet}
Olga Russakovsky, Jia Deng, Hao Su, Jonathan Krause, Sanjeev Satheesh, Sean Ma, Zhiheng Huang, Andrej Karpathy, Aditya Khosla, Michael Bernstein, Alexander~C Berg, and Li~Fei-Fei.
\newblock {ImageNet} {Large Scale Visual Recognition Challenge}.
\newblock \emph{International Journal of Computer Vision}, 115\penalty0 (3):\penalty0 211--252, 2015.

\bibitem[Sainath and Parada(2015)]{sainath2015cnn}
Tara~N. Sainath and Carolina Parada.
\newblock Convolutional neural networks for small-footprint keyword spotting.
\newblock In \emph{Interspeech}, pages 1478--1482, 2015.
\newblock \doi{10.21437/Interspeech.2015-352}.

\bibitem[Saito and Rehmsmeier(2015)]{saito2015prroc}
Takaya Saito and Marc Rehmsmeier.
\newblock The precision-recall plot is more informative than the {ROC} plot when evaluating binary classifiers on imbalanced datasets.
\newblock \emph{PLOS ONE}, 10\penalty0 (3):\penalty0 e0118432, 2015.
\newblock \doi{10.1371/journal.pone.0118432}.

\bibitem[Sakthi et~al.(2021)Sakthi, Desai, Hamilton, and Tewfik]{sakthi2021keyword}
Madhumitha Sakthi, Maansi Desai, Liberty Hamilton, and Ahmed Tewfik.
\newblock Keyword-spotting and speech onset detection in {EEG}-based brain computer interfaces.
\newblock In \emph{2021 10th International IEEE/EMBS Conference on Neural Engineering (NER)}, pages 519--522. IEEE, 2021.

\bibitem[Sato et~al.(2024)Sato, Tomeoka, Horiguchi, Arulkumaran, Kanai, and Sasai]{sato2024scaling}
Motoshige Sato, Kenichi Tomeoka, Ilya Horiguchi, Kai Arulkumaran, Ryota Kanai, and Shuntaro Sasai.
\newblock Scaling law in neural data: Non-invasive speech decoding with 175 hours of {EEG} data.
\newblock \emph{arXiv preprint arXiv:2407.07595}, 2024.

\bibitem[Schirrmeister et~al.(2017)Schirrmeister, Springenberg, Fiederer, Glasstetter, Eggensperger, Tangermann, Hutter, Burgard, and Ball]{schirrmeister2017deep}
Robin~Tibor Schirrmeister, Jost~Tobias Springenberg, Lukas Dominique~Josef Fiederer, Martin Glasstetter, Katharina Eggensperger, Michael Tangermann, Frank Hutter, Wolfram Burgard, and Tonio Ball.
\newblock Deep learning with convolutional neural networks for {EEG} decoding and visualization.
\newblock \emph{Human Brain Mapping}, 38\penalty0 (11):\penalty0 5391--5420, 2017.
\newblock \doi{10.1002/hbm.23730}.

\bibitem[Schoffelen et~al.(2019)]{schoffelen2019mous}
Jan-Mathijs Schoffelen et~al.
\newblock A 204-subject multimodal human neuroimaging dataset to study language processing.
\newblock \emph{Scientific Data}, 6\penalty0 (1):\penalty0 1--13, 2019.
\newblock \doi{10.1038/s41597-019-0020-y}.

\bibitem[Tang et~al.(2023)Tang, LeBel, Jain, and Huth]{tang2023semantic}
Jerry Tang, Alexander LeBel, Shailee Jain, and Alexander~G. Huth.
\newblock Semantic reconstruction of continuous language from non-invasive brain recordings.
\newblock \emph{Nature Neuroscience}, 26\penalty0 (5):\penalty0 858--866, 2023.
\newblock \doi{10.1038/s41593-023-01304-9}.

\bibitem[Tezcan et~al.(2023)Tezcan, Weissbart, and Martin]{tezcan2023tradeoff}
Filiz Tezcan, Hugo Weissbart, and Andrea~E. Martin.
\newblock A tradeoff between acoustic and linguistic feature encoding of continuous speech in the human brain.
\newblock \emph{eLife}, 12:\penalty0 e82386, 2023.
\newblock \doi{10.7554/eLife.82386}.

\bibitem[Vitevitch and Luce(1999)]{VitevitchLuce1999_JML}
Michael~S. Vitevitch and Paul~A. Luce.
\newblock Probabilistic phonotactics and neighborhood activation in spoken word recognition.
\newblock \emph{Journal of Memory and Language}, 40\penalty0 (3):\penalty0 374--408, 1999.
\newblock \doi{10.1006/jmla.1998.2618}.

\bibitem[Warden(2018)]{warden2018speechcommands}
Pete Warden.
\newblock Speech commands: A dataset for limited-vocabulary speech recognition.
\newblock \emph{arXiv preprint arXiv:1804.03209}, 2018.

\bibitem[Weide(1998)]{weide1998carnegie}
Robert Weide.
\newblock The {CMU} pronouncing dictionary, 1998.
\newblock URL \url{http://www.speech.cs.cmu.edu/cgi-bin/cmudict}.
\newblock release 0.6.

\bibitem[Willett et~al.(2023)Willett, Kunz, Fan, Avansino, Wilson, Choi, Kamdar, Glasser, Hochberg, Druckmann, Shenoy, and Henderson]{willett2023nature}
Francis~R. Willett, Erin~M. Kunz, Chaofei Fan, Donald~T. Avansino, Guy~H. Wilson, Eun~Young Choi, Foram Kamdar, Matthew~F. Glasser, Leigh~R. Hochberg, Shaul Druckmann, Krishna~V. Shenoy, and Jaimie~M. Henderson.
\newblock A high-performance speech neuroprosthesis.
\newblock \emph{Nature}, 620\penalty0 (7976):\penalty0 1031--1036, 2023.
\newblock \doi{10.1038/s41586-023-06377-x}.

\bibitem[Yang et~al.(2024{\natexlab{a}})Yang, Duan, Jo, Zhang, Xu, Jones, Hu, Lin, and Xiong]{yang2024neugpt}
Yiqian Yang, Yiqun Duan, Hyejeong Jo, Qiang Zhang, Renjing Xu, Oiwi~Parker Jones, Xuming Hu, Chin-teng Lin, and Hui Xiong.
\newblock {NeuGPT}: Unified multi-modal neural {GPT}.
\newblock \emph{arXiv preprint arXiv:2410.20916}, 2024{\natexlab{a}}.

\bibitem[Yang et~al.(2024{\natexlab{b}})Yang, Duan, Zhang, Xu, and Xiong]{yang2024neuspeech}
Yiqian Yang, Yiqun Duan, Qiang Zhang, Renjing Xu, and Hui Xiong.
\newblock {NeuSpeech}: Decode neural signal as speech.
\newblock \emph{arXiv preprint}, 2024{\natexlab{b}}.
\newblock URL \url{https://arxiv.org/abs/2403.01748}.

\bibitem[Yang et~al.(2024{\natexlab{c}})Yang, Jo, Duan, Zhang, Zhou, Lee, Xu, and Xiong]{yang2024mad}
Yiqian Yang, Hyejeong Jo, Yiqun Duan, Qiang Zhang, Jinni Zhou, Won~Hee Lee, Renjing Xu, and Hui Xiong.
\newblock {MAD}: Multi-alignment {MEG}-to-text decoding.
\newblock \emph{arXiv preprint arXiv:2406.01512}, 2024{\natexlab{c}}.

\bibitem[Yue et~al.(2007)Yue, Finley, Radlinski, and Joachims]{yue2007ap}
Yisong Yue, Thomas Finley, Filip Radlinski, and Thorsten Joachims.
\newblock A support vector method for optimizing average precision.
\newblock In \emph{Proceedings of the 30th Annual International ACM SIGIR Conference on Research and Development in Information Retrieval}, pages 271--278, Amsterdam, The Netherlands, 2007. ACM.
\newblock \doi{10.1145/1277741.1277790}.

\bibitem[Zhang et~al.(2020)Zhang, Kishore, Wu, Weinberger, and Artzi]{bertscore}
Tianyi Zhang, Varsha Kishore, Felix Wu, Kilian~Q. Weinberger, and Yoav Artzi.
\newblock Bertscore: Evaluating text generation with bert.
\newblock \emph{International Conference on Learning Representations (ICLR)}, 2020.
\newblock URL \url{https://openreview.net/forum?id=SkeHuCVFDr}.

\end{thebibliography}


\appendix

\section{Open Resources: Code, Tutorial, and Leaderboard}\label{appendix:code}
\subsection{Updated \texttt{pnpl} datasets}
We release an updated version of the open source \texttt{pnpl} library (\cite{ozdogan2025dataset}) to support word-level tasks. This allows both full signal-to-word and single/multi-keyword tasks to be performed using similar syntax to the existing \texttt{LibriBrainSpeech} and \texttt{LibriBrainPhoneme} classes:

\textbf{Word-level task:}
\begin{verbatim}
    from pnpl.datasets import LibriBrainWord
    
    dataset = LibriBrainWord(
        data_path="./data/",
        partition="train",
        tmin=0.0,
        tmax=0.8,
    )
\end{verbatim}

\textbf{Single-keyword task:}
\begin{verbatim}
from pnpl.datasets import LibriBrainWord

dataset = LibriBrainWord(
    data_path="./data/",
    partition="train",
    keyword_detection="watson",
)
\end{verbatim}

\textbf{Multi-keyword task:}
\begin{verbatim}
from pnpl.datasets import LibriBrainWord

dataset = LibriBrainWord(
    data_path="./data/",
    partition="train",
    keyword_detection=["sherlock", "holmes"],
)
\end{verbatim}

For the single- or multi-keyword task, sample length is inferred from the longest keyword duration and can be extended with the \texttt{positive\_buffer} and \texttt{negative\_buffer} arguments. Overwrites using \texttt{tmin} and \texttt{tmax} are of course possible. For full signal-to-word, that is rarely the intended behaviour, so these options are disabled and a reasonable default is used instead.

Similarly, the \texttt{keyword\_detection} variant will verify that the keyword(s) are present in the dataset and, if not, default to highest prevalence sessions as validation and test sets, while the signal-to-word variant will use the default validation and test sets.

The library is available on PyPI\footnote{https://pypi.org/project/pnpl/} and on GitHub\footnote{https://github.com/neural-processing-lab/pnpl}.

\subsection{Tutorial Notebook}
To encourage further exploration within the community, we also release a tutorial in the format of a Jupyter Notebook. Within the compute limits of the Colab Free Tier (\texttt{T4} GPU), the notebook allows for training a model around 10\% of the LibriBrain dataset, reaching significantly above chance performance in under 30 minutes.
The notebook is available in the \texttt{tutorial} folder of the \texttt{keyword-experiments} repository\footnote{http://github.com/neural-processing-lab/libribrain-keyword-experiments}.

\subsection{Experiment Code}
Finally, to allow for full reproducibility, we release the code for the experiments and analysis conducted in this paper. The code is available in the \texttt{experiments} folder of the \texttt{keyword-experiments} repository\footnote{http://github.com/neural-processing-lab/libribrain-keyword-experiments}.

\section{Dataset Figures}

\begin{figure}[ht]
    \centering
    \includegraphics[width=0.95\textwidth]{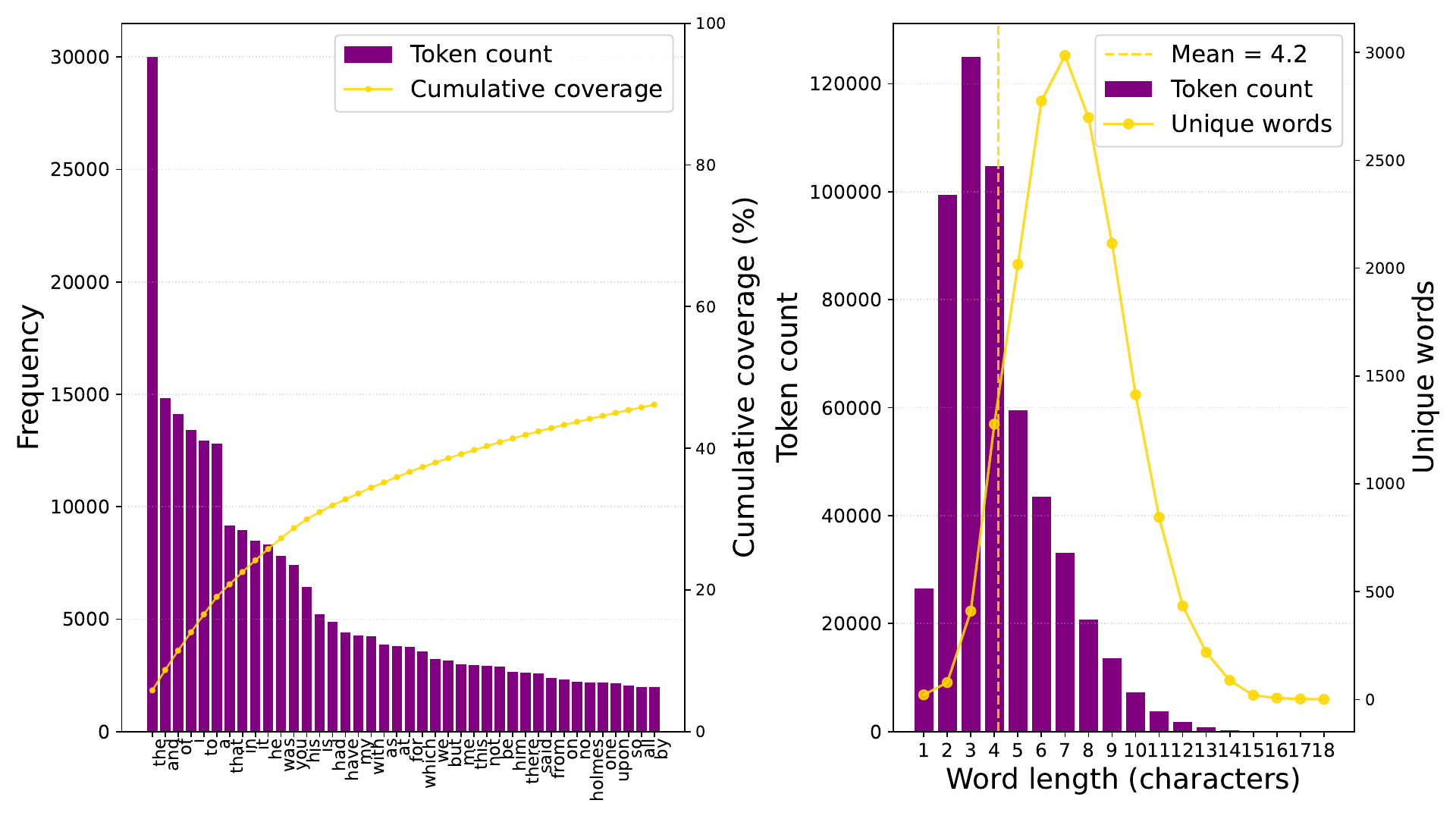}
    \caption{Overview of the LibriBrain dataset. (A) 40 most common words and their coverage of the dataset. (B) Word length distribution and the number of unique words for each length.}
    \label{fig:libribrain-overview}
\end{figure}

\begin{figure}[ht]
    \centering
    \includegraphics[width=0.95\textwidth]{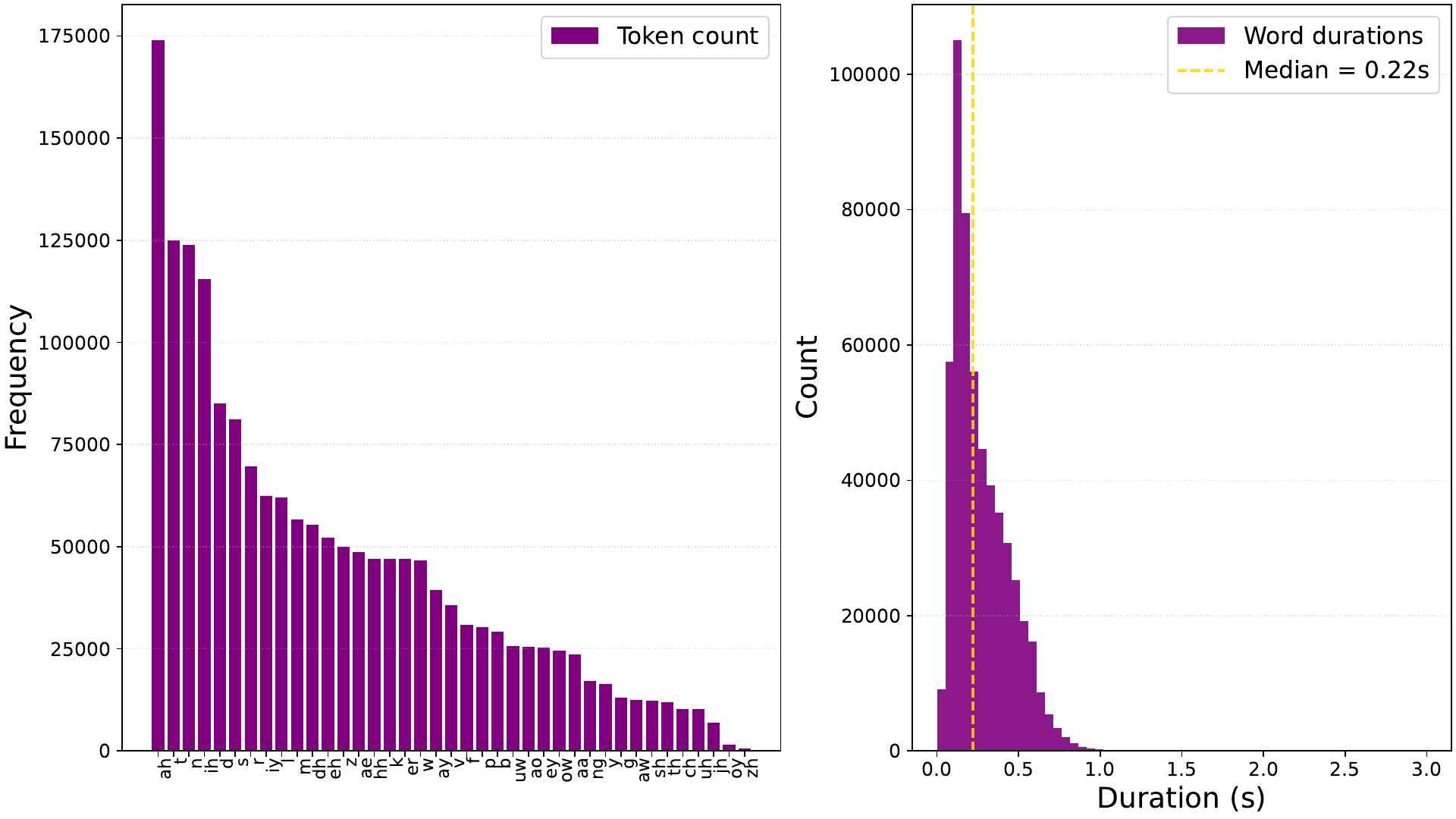}
    \caption{(C) Phoneme distribution across the corpus (39 ARPAbet phonemes from \citep{weide1998carnegie}). (D) Word duration distribution with median line.}
    \label{fig:libribrain-phoneme-duration}
\end{figure}

\section{Dataset Tables}

\subsection{Data Scaling}
\begin{table}[!htbp]
\centering
\small
\begin{tabular}{lcccc}
\hline
\textbf{Training fraction} & \textbf{AUPRC} (\(\pm\) SE) & \textbf{AUROC} (\(\pm\) SE) & \textbf{AUPRC p-value} & \textbf{AUPRC/base rate (\(\times\))} \\
\hline
5\%   & 0.009 \( \pm \) 0.003 & 0.626 \( \pm \) 0.058 & 0.108         & 1.69 \\
10\%  & 0.019 \( \pm \) 0.009 & 0.746 \( \pm \) 0.049 & 0.0156        & 3.67 \\
20\%  & 0.027 \( \pm \) 0.020 & 0.733 \( \pm \) 0.050 & $6.0\times10^{-4}$ & 5.18 \\
40\%  & 0.044 \( \pm \) 0.027 & 0.782 \( \pm \) 0.046 & $<5\times10^{-5}$ & 8.59 \\
60\%  & 0.032 \( \pm \) 0.016 & 0.784 \( \pm \) 0.046 & $2.0\times10^{-4}$ & 6.28 \\
80\%  & 0.048 \( \pm \) 0.029 & 0.796 \( \pm \) 0.047 & $<5\times10^{-5}$ & 9.37 \\
100\% & 0.045 \( \pm \) 0.021 & 0.834 \( \pm \) 0.037 & $<5\times10^{-5}$ & 8.66 \\
\hline
\end{tabular}
\caption{Detailed scaling results for keyword detection across training fractions (seed-averaged over three runs). Standard errors are approximated from 95\% bootstrap CIs as \(\mathrm{SE} \approx (\mathrm{CI}_{\mathrm{hi}} - \mathrm{CI}_{\mathrm{lo}})/3.92\). P-values are from one-sided permutation tests of the seed-average AUPRC against the null. The base rate for the fixed test set is 0.00515.}
\label{tab:scaling-detailed}
\end{table}

\FloatBarrier

\subsection{Keyword Choice}
\begin{table}[!htbp]
\centering
\small
\begin{tabular}{lccccc}
\hline
\textbf{Keyword} & \textbf{Base rate} & \textbf{AUPRC} & \textbf{AUROC} & \textbf{Acc} & \textbf{Best F1} \\
\hline
and & 0.039 & 0.218 \( \pm \) 0.014 & \textbf{0.825} \( \pm \) 0.002 & 0.756 \( \pm \) 0.010 & \textbf{0.292} \( \pm \) 0.008 \\
the & 0.077 & 0.213 \( \pm \) 0.005 & 0.728 \( \pm \) 0.006 & 0.673 \( \pm \) 0.004 & 0.278 \( \pm \) 0.006 \\
i & 0.039 & 0.191 \( \pm \) 0.005 & 0.784 \( \pm \) 0.004 & 0.708 \( \pm \) 0.015 & 0.279 \( \pm \) 0.006 \\
watson & 0.005 & \textbf{0.065} \( \pm \) 0.017 & 0.759 \( \pm \) 0.006 & 0.952 \( \pm \) 0.034 & 0.149 \( \pm \) 0.036 \\
holmes & 0.008 & 0.028 \( \pm \) 0.001 & 0.791 \( \pm \) 0.013 & 0.820 \( \pm \) 0.066 & 0.072 \( \pm \) 0.005 \\
himself & 0.003 & 0.013 \( \pm \) 0.001 & 0.758 \( \pm \) 0.021 & 0.993 \( \pm \) 0.001 & 0.062 \( \pm \) 0.008 \\
considerable & 0.001 & 0.012 \( \pm \) 0.007 & 0.684 \( \pm \) 0.066 & 0.997 \( \pm \) 0.002 & 0.041 \( \pm \) 0.025 \\
inspector & 0.004 & 0.008 \( \pm \) 0.001 & 0.593 \( \pm \) 0.020 & 0.994 \( \pm \) 0.001 & 0.040 \( \pm \) 0.012 \\
mister & 0.002 & 0.007 \( \pm \) 0.004 & 0.505 \( \pm \) 0.051 & 0.998 \( \pm \) 0.000 & 0.054 \( \pm \) 0.022 \\
remarkable & 0.001 & 0.003 \( \pm \) 0.001 & 0.658 \( \pm \) 0.070 & 0.991 \( \pm \) 0.008 & 0.010 \( \pm \) 0.004 \\
understand & 0.001 & 0.002 \( \pm \) 0.001 & 0.523 \( \pm \) 0.136 & \textbf{0.999} \( \pm \) 0.000 & 0.006 \( \pm \) 0.003 \\
investigation & 0.001 & 0.002 \( \pm \) 0.001 & 0.665 \( \pm \) 0.047 & 0.998 \( \pm \) 0.001 & 0.011 \( \pm \) 0.007 \\
\hline
\end{tabular}
\caption{Seed-averaged per-keyword metrics (absolute units): base rate (positive prevalence), AUPRC, AUROC, Accuracy, and Best F1 (per-seed best across thresholds). Means and $\pm$ SEM are computed across seeds. Bold marks the best improvement vs base rate for AUPRC, and the highest mean for other columns.}
\label{tab:keyword-metrics}
\end{table}

\FloatBarrier

\subsection{Operating Points}
\begin{table}[!htbp]
\centering
\small
\begin{tabular}{lccc}
\hline
\textbf{Scenario} & \textbf{Metric} & \textbf{Value} & \textbf{SE} \\
\hline
Assistive (\(\lambda=2\)/h), target recall \(\approx 0.10\) & FA/h & 2.194 & 1.629 \\
Assistive (\(\lambda=2\)/h), FA/h budget 2.0 & Recall & 0.139 & 0.050 \\
Assistive (\(\lambda=2\)/h), FA/h budget 0.5 & Recall & 0.083 & 0.024 \\
Labelled test coverage & FP/h & 16.3 & 12.1 \\
\hline
\end{tabular}
\caption{Operating-point snapshot (best-AUPRC buffer: neg=0.1s, pos=0.3s). Values are seed-averages $\pm$ SE (\(n=3\)). Scenario-scale metrics use the assistive case (\(\lambda=2\)/h).}
\label{tab:operating-points}
\end{table}

\FloatBarrier

\subsection{Keyword Length (matched frequency)}
\begin{table}[!htbp]
\centering
\small
\begin{tabular}{lcccc}
\hline
\textbf{Keyword} & \textbf{Chars} & \textbf{Base rate} & \textbf{AUPRC} & \textbf{\%$\Delta$AUPRC over base} \\
\hline
walk & 4 & 0.00056 & 0.00136 $\pm$ 0.00010 & 144.9 $\pm$ 18.0 \\
surely & 6 & 0.00056 & 0.00141 $\pm$ 0.00031 & 153.7 $\pm$ 59.3 \\
excellent & 9 & 0.00056 & 0.00133 $\pm$ 0.00033 & 138.9 $\pm$ 57.6 \\
\hline
\end{tabular}
\caption{Matched-frequency keyword comparison (seed-averaged over three runs). Results show no significant differences (overlapping SEMs).}
\label{tab:keyword-length-matched}
\end{table}

\FloatBarrier

\subsection{Temporal Offsets}
\begin{table}[!htbp]
\centering
\small
\begin{tabular}{r r c c r}
\hline
 \textbf{Neg [s]} & \textbf{Pos [s]} & \textbf{AUPRC (mean $\pm$ SE)} & \textbf{AUROC (mean $\pm$ SE)} & \textbf{Seeds} \\
\hline
 0.00 & 0.00 & $ 0.039 \pm 0.009 $ & $ 0.811 \pm 0.011 $ & 3 \\
 0.00 & 0.05 & $ 0.043 \pm 0.003 $ & $ 0.813 \pm 0.008 $ & 3 \\
 0.00 & 0.10 & $ 0.030 \pm 0.007 $ & $ 0.779 \pm 0.015 $ & 3 \\
 0.00 & 0.15 & $ 0.039 \pm 0.009 $ & $ 0.799 \pm 0.021 $ & 3 \\
 0.00 & 0.20 & $ 0.037 \pm 0.003 $ & $ 0.781 \pm 0.011 $ & 3 \\
 0.00 & 0.25 & $ 0.064 \pm 0.007 $ & $ 0.820 \pm 0.002 $ & 3 \\
 0.00 & 0.30 & $ 0.052 \pm 0.006 $ & $ 0.821 \pm 0.007 $ & 3 \\
 0.05 & 0.00 & $ 0.083 \pm 0.029 $ & $ 0.799 \pm 0.025 $ & 3 \\
 0.05 & 0.05 & $ 0.049 \pm 0.010 $ & $ 0.781 \pm 0.010 $ & 3 \\
 0.05 & 0.10 & $ 0.039 \pm 0.006 $ & $ 0.780 \pm 0.013 $ & 3 \\
 0.05 & 0.15 & $ 0.040 \pm 0.013 $ & $ 0.758 \pm 0.002 $ & 3 \\
 0.05 & 0.20 & $ 0.053 \pm 0.007 $ & $ 0.812 \pm 0.009 $ & 3 \\
 0.05 & 0.25 & $ 0.069 \pm 0.021 $ & $ 0.800 \pm 0.010 $ & 3 \\
 0.05 & 0.30 & $ 0.045 \pm 0.014 $ & $ 0.796 \pm 0.003 $ & 3 \\
 0.10 & 0.00 & $ 0.025 \pm 0.005 $ & $ 0.730 \pm 0.011 $ & 3 \\
 0.10 & 0.05 & $ 0.069 \pm 0.031 $ & $ 0.826 \pm 0.005 $ & 3 \\
 0.10 & 0.10 & $ 0.047 \pm 0.011 $ & $ 0.789 \pm 0.005 $ & 3 \\
 0.10 & 0.15 & $ 0.070 \pm 0.003 $ & $ 0.787 \pm 0.020 $ & 3 \\
 0.10 & 0.20 & $ 0.080 \pm 0.029 $ & $ 0.836 \pm 0.006 $ & 3 \\
 0.10 & 0.25 & $ 0.071 \pm 0.007 $ & $ 0.787 \pm 0.014 $ & 3 \\
 \textbf{0.10} & \textbf{0.30} & $ \mathbf{0.094} \pm \mathbf{0.032} $ & $ \mathbf{0.804} \pm \mathbf{0.017} $ & \textbf{3} \\
 0.15 & 0.00 & $ 0.038 \pm 0.012 $ & $ 0.764 \pm 0.027 $ & 3 \\
 0.15 & 0.05 & $ 0.027 \pm 0.005 $ & $ 0.781 \pm 0.028 $ & 3 \\
 0.15 & 0.10 & $ 0.053 \pm 0.014 $ & $ 0.760 \pm 0.010 $ & 3 \\
 0.15 & 0.15 & $ 0.036 \pm 0.003 $ & $ 0.779 \pm 0.013 $ & 3 \\
 0.15 & 0.20 & $ 0.030 \pm 0.003 $ & $ 0.795 \pm 0.014 $ & 3 \\
 0.15 & 0.25 & $ 0.027 \pm 0.006 $ & $ 0.789 \pm 0.017 $ & 3 \\
 0.15 & 0.30 & $ 0.034 \pm 0.007 $ & $ 0.797 \pm 0.017 $ & 3 \\
 0.20 & 0.00 & $ 0.029 \pm 0.000 $ & $ 0.764 \pm 0.000 $ & 1 \\
 0.20 & 0.05 & $ 0.045 \pm 0.010 $ & $ 0.819 \pm 0.009 $ & 3 \\
 0.20 & 0.10 & $ 0.046 \pm 0.015 $ & $ 0.810 \pm 0.004 $ & 3 \\
 0.20 & 0.15 & $ 0.036 \pm 0.003 $ & $ 0.801 \pm 0.020 $ & 3 \\
 0.20 & 0.20 & $ 0.042 \pm 0.003 $ & $ 0.792 \pm 0.010 $ & 3 \\
 0.20 & 0.25 & $ 0.037 \pm 0.009 $ & $ 0.822 \pm 0.011 $ & 3 \\
 0.20 & 0.30 & $ 0.050 \pm 0.012 $ & $ 0.836 \pm 0.008 $ & 3 \\
\hline
\end{tabular}
\caption{Seed-averaged performance across temporal offsets around the keyword onset. Values are mean $\pm$ standard error across seeds. The row with the highest \%$\Delta$AUPRC over the 0/0 baseline is typeset in bold.}
\label{tab:buffer-offsets-full}
\end{table}



\end{document}